\PassOptionsToPackage{table}{xcolor}
\documentclass{article}

\usepackage[eandd, final]{neurips_2026}

\usepackage[utf8]{inputenc} 
\usepackage[T1]{fontenc}    
\usepackage{hyperref}       
\usepackage{url}            
\usepackage{booktabs}       
\usepackage{amsfonts}       
\usepackage{nicefrac}       
\usepackage{microtype}      
\usepackage{xcolor}         

\usepackage{amsmath,amssymb,graphicx,array,tabularx,placeins,tikz}
\usepackage{needspace}
\usetikzlibrary{patterns}
\newcommand{\removalpool}{Removal}
\newcommand{\sourcepool}{Source-only}

\newcommand{\RemovalDelta}{-0.2132}
\newcommand{\RemovalCI}{[-0.2200,-0.2065]}

\newcommand{\FiveRowDelta}{+0.0011}
\newcommand{\FiveRowCI}{[-0.0044,+0.0066]}
\newcommand{\InternalMainDelta}{+0.0051}
\newcommand{\InternalMainCI}{[-0.0043,+0.0145]}
\newcommand{\InternalSourceDelta}{+0.0037}
\newcommand{\InternalSourceCI}{[-0.0049,+0.0122]}
\newcommand{\FreshBCompCI}{[+0.0045,+0.0140]}
\newcommand{\KVShare}{71.8}
\newcommand{\PhashRate}{11.92}
\newcommand{\ChosenNCCRate}{4.66}
\newcommand{\AnyNCCRate}{6.90}

\newcommand{\TrainedMSeven}{-0.062}
\newcommand{\ProbDropPP}{0.355}
\newcommand{\ProbCILoPP}{-0.486}
\newcommand{\ProbCIHiPP}{-0.224}
\newcommand{\RemovalNonKVDeltaPP}{1.38}
\newcommand{\CMinusAPercentileCI}{[-0.0077,-0.0006]}
\newcommand{\ResNetPublicCI}{[-0.2024,-0.1979]}
\newcommand{\ResNetNonKVCI}{[+0.0836,+0.0921]}
\newcommand{\ConvNeXtPublicCI}{[-0.2137,-0.2055]}
\newcommand{\ConvNeXtNonKVCI}{[+0.0919,+0.0990]}

\newcommand{\TrainingFullAUC}{0.9879}
\newcommand{\TrainingRemovalAUC}{0.8812}
\newcommand{\TrainingFullCombined}{0.9066}
\newcommand{\TrainingRemovalCombined}{0.7466}
\newcommand{\TrainingCombinedDelta}{-0.1600}

\newcommand{\TrainingFullBestBA}{0.8306}
\newcommand{\TrainingRemovalBestBA}{0.6178}
\newcommand{\TrainingBestDelta}{-0.2129}
\newcommand{\TrainingFinalDelta}{-0.2132}
\newcommand{\TrainingAddbackBA}{0.7048}
\newcommand{\TrainingAddbackComparatorBA}{0.6101}
\newcommand{\TrainingAddbackDelta}{+0.0947}

\newcommand{\BalancedCoarseningMaxShiftPP}{0.27}

\newcommand{\BalancedCoarseningFortyOneRange}{[6.899, 17.996]}
\newcommand{\BalancedCoarseningThirtyRange}{[6.858, 18.014]}
\newcommand{\BalancedCoarseningTwentyRange}{[6.888, 17.975]}
\newcommand{\BalancedCoarseningTenRange}{[6.925, 17.729]}

\newcommand{\ScoreConsistentPooledNonKV}{-0.050}
\newcommand{\ScoreConsistentEqualNonKV}{-0.020}
\newcommand{\ScoreConsistentPublicRho}{0.557}
\newcommand{\ScoreConsistentNonKVRho}{0.459}

\newcommand{\KvasirAuxRuns}{28}
\newcommand{\KvasirDiversityHighTerminal}{0.240}
\newcommand{\KvasirDiversityLowTerminal}{0.184}
\newcommand{\KvasirGroupTestClassMax}{9}
\newcommand{\KvasirGroupTestClassMin}{6}
\newcommand{\KvasirScaleFullTerminal}{0.234}
\newcommand{\KvasirScaleHalfTerminal}{0.236}

\newcommand{\LocalAccuracyMaxPercent}{99.98}
\newcommand{\LocalAccuracyMinPercent}{99.56}
\newcommand{\LocalExposedProbabilityPP}{-0.321}

\newcommand{\LocalProbabilityLOPOMaxPP}{-0.215}
\newcommand{\LocalProbabilityLOPOMinPP}{-0.390}
\newcommand{\LocalTopThreePrefixSharePercent}{85.0}
\newcommand{\LocalUnflaggedProbabilityPP}{+0.034}

\hypersetup{
  hidelinks,
  pdftitle={ReuScope: Auditing Data Reuse and Evaluation Sensitivity in Capsule Vision 2024},
  pdfauthor={Eunseob Choi; Kyeonghun Kim; Hyuk-Jae Lee; Nam-Joon Kim},
  pdfsubject={NeurIPS 2026 Evaluations and Datasets Track},
  pdfkeywords={ReuScope, benchmark auditing, capsule endoscopy, data reuse, evaluation sensitivity}
}

\title{ReuScope: Auditing Data Reuse and Evaluation Sensitivity in Capsule Vision 2024}
\author{%
  \textbf{Eunseob Choi}\textsuperscript{1}\quad
  \textbf{Kyeonghun Kim}\textsuperscript{2}\quad
  \textbf{Hyuk-Jae Lee}\textsuperscript{3}\quad
  \textbf{Nam-Joon Kim}\textsuperscript{3}\thanks{Corresponding author.}\\[5pt]
  {\small\textsuperscript{1}Gwangju Institute of Science and Technology}\\[1pt]
  {\small\textsuperscript{2}OUTTA\qquad\textsuperscript{3}Seoul National University}\\[5pt]
  {\footnotesize\texttt{eunseobchoi@gm.gist.ac.kr}\quad\texttt{kyeonghun.kim@outta.ai}}\\[-1pt]
  {\footnotesize\texttt{hjlee@capp.snu.ac.kr}\quad\texttt{knj01@snu.ac.kr}}
}
\definecolor{TableHeader}{HTML}{F2F2F2}
\definecolor{ClaimTint}{HTML}{E8F0F5}
\newcommand{\PaperRepo}{ReuScope}
\definecolor{PaperLink}{HTML}{244A68}
\hypersetup{colorlinks=true,linkcolor=black,citecolor=black,urlcolor=PaperLink,
  bookmarksopen=true,bookmarksnumbered=true}
\makeatletter
\renewcommand{\@maketitle}{%
  \vbox{\hsize\textwidth\linewidth\hsize
    \vbox to 0pt{%
      \vss
      \hbox to \textwidth{%
        \raisebox{-.5\height}{\includegraphics[height=4.5mm]{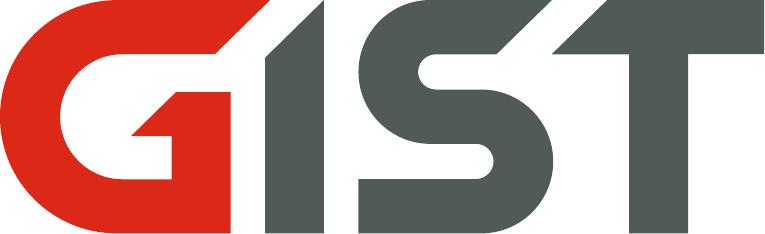}}%
        \hspace{6mm}%
        \raisebox{-.5\height}{\includegraphics[height=8mm]{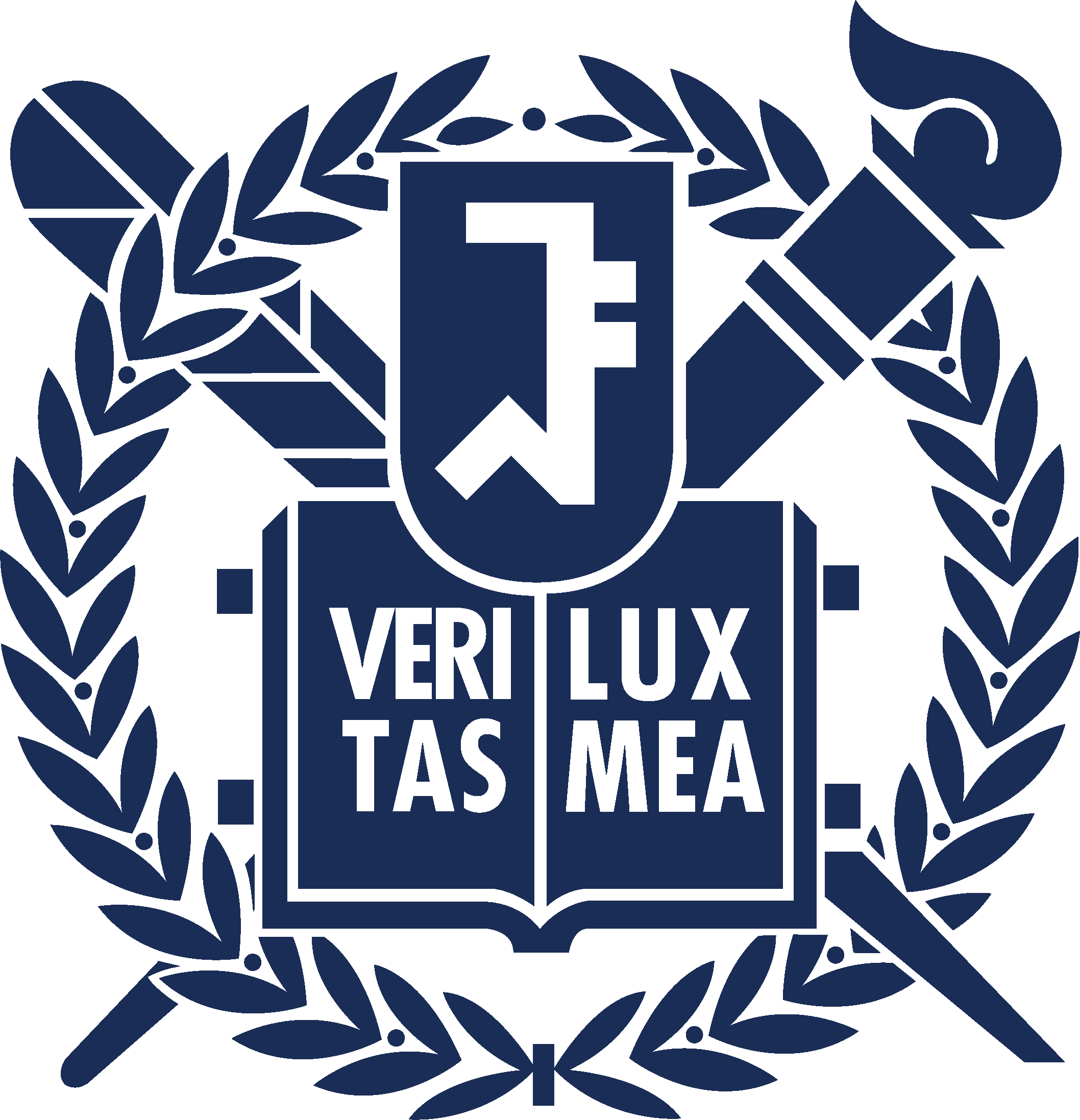}}%
        \hfil
      }%
      \vskip 23pt
    }%
    \vskip 2pt
    \centering
    {\fontsize{17}{20}\selectfont\bfseries\@title\par}
    \vskip 13pt
    {\normalsize\@author\par}
    \vskip 10pt
    {\fontsize{9}{11}\selectfont
      \textbf{Code \& Data:}\enspace
      \href{https://github.com/eunseobchoi/\PaperRepo}{\texttt{github.com/eunseobchoi/\PaperRepo}}\par}
    \vskip 10pt
    {\color{black!25}\hrule height .4pt}
    \vskip 6pt
  }%
}
\long\def\@makecaption#1#2{%
  \vskip\abovecaptionskip
  \sbox\@tempboxa{\small\textbf{#1.}\enspace #2}%
  \ifdim\wd\@tempboxa>\hsize
    {\small\textbf{#1.}\enspace #2\par}%
  \else
    \global\@minipagefalse
    \hb@xt@\hsize{\hfil\box\@tempboxa\hfil}%
  \fi
  \vskip\belowcaptionskip
}
\makeatother

\begin{document}
\addtocontents{toc}{\protect\setcounter{tocdepth}{-1}}
\maketitle
\begin{abstract}
We develop ReuScope, a protocol for auditing data reuse and evaluation sensitivity, using Capsule Vision 2024 as a case study. It combines measurements of data relationships with two comparisons: changing the evaluation population while holding predictions fixed, and changing training data while holding evaluation fixed. Byte-identical files cross the public training--validation boundary, and all 41 Kvasir-Capsule recordings that contribute frames occur in both partitions. Shared source recordings establish cross-split overlap; missing cross-recording patient identifiers prevent counting unique shared patients. Across ten paired seeds, removing all training rows attributed to Kvasir-Capsule, plus five rows from other sources that pass a hash screen against it, lowers public-validation balanced accuracy from 82.5\% to 61.2\%. Evaluating the same fitted models on the validation rows of the other sources reverses the comparison: removal improves balanced accuracy by $\RemovalNonKVDeltaPP$ percentage points. The public-validation drop measures sensitivity to source removal, not an amount of leakage-induced inflation. Matched controls show that sample size and class counts alone do not explain the observed score differences. An exploratory experiment first removes the close training matches of flagged validation frames, then also removes either the neighbors of those matches or equally many distant training rows. Relative to matched unflagged frames, the difference-in-differences estimate in true-class probability is $-\ProbDropPP$ percentage points; the paired-$t$ interval for accuracy spans zero. The resulting procedure distinguishes observed reuse from conditional performance sensitivity and specifies the population, class support and uncertainty unit for each conclusion.
\end{abstract}

\section{Introduction}
\label{sec:intro}
A new filename does not imply new image content, a new source video or a new patient. In capsule endoscopy, neighboring frames can depict nearly the same scene, and identical files can occur under different names or labels. These relationships matter when public-validation scores are interpreted as evidence of generalization. Detecting reuse and identifying its effect on a score are nevertheless different tasks~\citep{kapoor2023leakage,how_you_split_2023}. Removing a source can reduce such reuse, but it also removes useful content and changes class support (rows per class), sample size and update count.

Capsule Vision 2024 (CV2024) was designed to support clinical frame review through ten-class abnormality classification and to encourage vendor-independent models~\citep{capsule_vision_2024}. Its public release combines four acquisition sources and shows exact file reuse, shared videos, disagreement between released labels and source-dependent performance. Beyond the challenge, EndoExtend24 reports organizer-evaluated CV2024 public-validation and official-test results~\citep{endoextend2024}. We ask whether the score change from removing a source keeps its direction across evaluation populations, and which matched controls are needed to interpret it.

We contribute the empirical CV2024 audit and ReuScope, a comparison and reporting protocol derived from its findings. The audit distinguishes data relationships, tests whether training contrasts persist across evaluation populations, and relates each conclusion to its population and uncertainty. Sections~\ref{sec:design}--\ref{sec:findings} present the comparisons and evidence; Section~\ref{sec:protocol} gives the resulting protocol.

\paragraph{Relationship to prior work.}
Leakage taxonomies and model information sheets relate data dependence to the population supporting a claim~\citep{kapoor2023leakage}; REFORMS extends this discipline to study design, uncertainty and reporting~\citep{kapoor2024reforms}. Visual audits characterize leakage by modality, coverage and degree~\citep{ramos2025dataleakage}, while ciFAIR connects detection, adjudication, test-set replacement and reevaluation~\citep{ciFAIR2020}. Medical-image studies benchmark duplicate detection~\citep{rauber2023medicaldedup} and document repeated content and lost provenance across derivative datasets~\citep{copycats_2024}. In endoscopy, EndoExtend24 propagates its revised 80/20 CV2024 split assignments to overlapping source images by filename~\citep{endoextend2024}; content and video-prefix checks reveal relationships that filename matching alone cannot establish. Our audit makes matching rules, training changes and evaluation populations explicit and separates observed reuse from the score response to a specified change.

\section{Case study and comparison design}
\label{sec:design}
\paragraph{Release and evaluation boundary.}
CV2024 contains 37,607 public training and 16,132 validation rows. The organizers describe a random 70--30 split and duplicate exclusion. They designate public validation as internal assessment and use a separate 4,385-image AIIMS test for external assessment and final ranking. Test images were available for prediction, but their labels were withheld until after evaluation. Final-test models could be trained on both public partitions~\citep{capsule_vision_2024,capsule_vision_trainval_figshare,capsule_vision_test_figshare}. We primarily audit the public training--validation boundary and analyze the official test separately.

By the organizers' source labels, validation contains 11,581 KVASIR (frames from Kvasir-Capsule), 4,289 SEE-AI, 165 KID and 97 AIIMS rows; KVASIR contributes $\KVShare\%$ of rows. Source totals match the challenge paper's Table I; the other sources are the SEE-AI project~\citep{yokote2024seeai}, KID~\citep{koulaouzidis2017kid} and an AIIMS collection~\citep{goel2022dilated}. Forty-one of the 43 labeled Kvasir-Capsule reference videos contribute to CV2024~\citep{smedsrud2021}. Their filename prefixes identify source recordings; patient identity across recordings is unavailable. We preserve full paths, labels and row multiplicities throughout.

\begin{figure}[htbp]
  \centering
  \includegraphics[width=\linewidth]{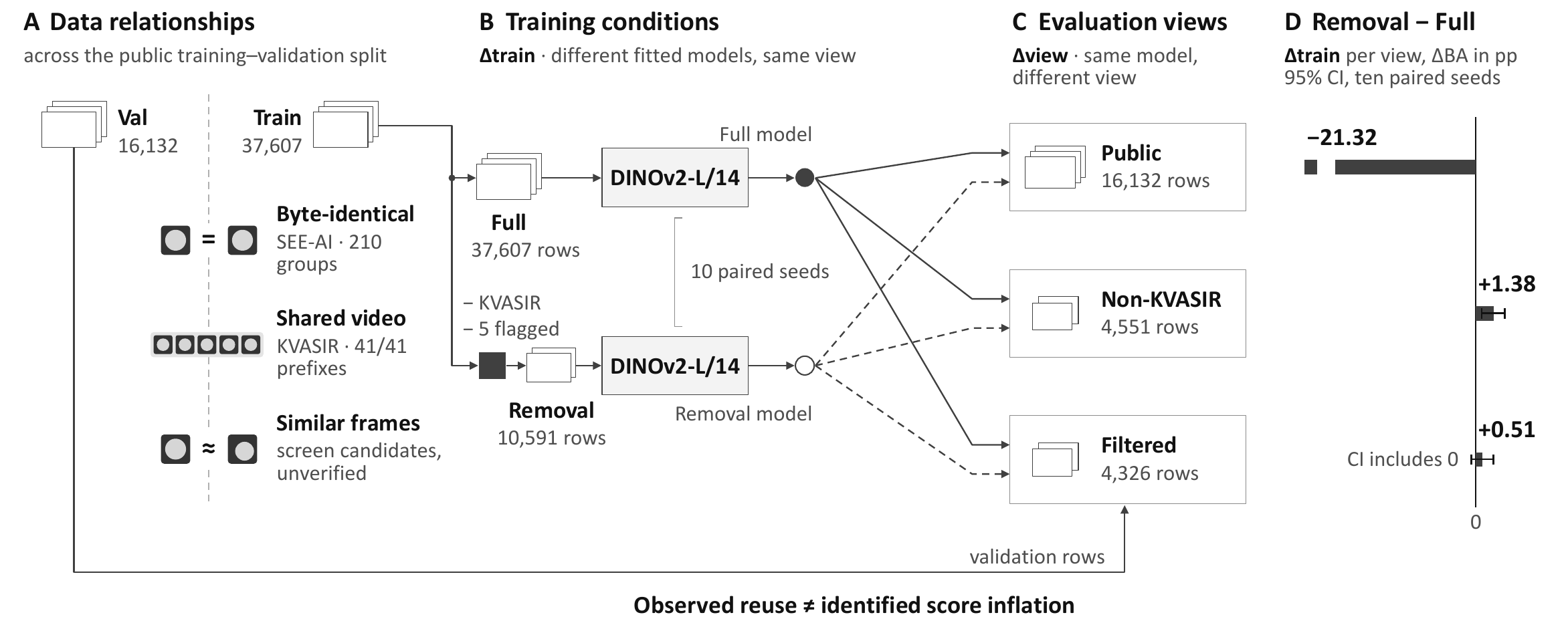}
  \caption{Data relationships and training comparisons across evaluation views in ReuScope. The protocol audits file identity, shared videos and similarity candidates across data splits. Paired Full and Removal models are evaluated on three validation populations, which separates training changes from evaluation changes. These performance contrasts show sensitivity to the evaluation population; they do not establish score inflation from data reuse.}
  \label{fig:protocol}
\end{figure}

\paragraph{Two comparisons.}
Figure~\ref{fig:protocol} illustrates why training and evaluation changes must be separated. Let $\mathcal{A}(T;s)$ denote the model produced by a specified training pipeline on list $T$ with seed $s$, and let $M_C(f,V)$ evaluate model $f$ on list $V$. The metric definition $C$ fixes the class-selection and aggregation rules. We distinguish
\begin{align}
\Delta_{\mathrm{view}}(f)&=M_C(f,V_1)-M_C(f,V_0),\label{eq:view}\\
\Delta_{\mathrm{train}}(V)&=\mathbb{E}_{s}\!\left[M_C(\mathcal{A}(T_1;s),V)-M_C(\mathcal{A}(T_0;s),V)\right].\label{eq:train}
\end{align}
The first changes evaluation with the model fixed; the second changes training with evaluation fixed and seeds paired. We call training lists pools and evaluation lists views. Crossing pools with views shows how training contrasts depend on the evaluation population.

\paragraph{Relationship measurements.}
We compute SHA-256 for every released public image and compare the bytes within each matching group. We also use 64-bit pHash and dHash descriptors~\citep{imagehash}; $d_p$ and $d_d$ denote their Hamming distances. The \emph{hash screen} against Kvasir-Capsule flags an image whose minimum $d_p$ and minimum $d_d$ over all reference images are both $\leq6$; the two minima may come from different references. The \emph{joint rule} instead requires a single matching image with $d_p\leq6$ and $d_d\leq6$, and the \emph{sum rule} one with $d_p+d_d\leq6$. We measure pixel similarity with normalized cross-correlation (NCC) and structural similarity (SSIM)~\citep{wang2004image}, after direct resizing to $96\times96$ grayscale. The thresholds specify screening rules, not calibrated duplicate classifiers.

\paragraph{Training controls and uncertainty.}
The primary pipeline uses DINOv2-L/14~\citep{dinov2_2023} with a 256-unit nonlinear head and rank-eight LoRA~\citep{hu2021lora} on the final four blocks. The primary conditions use ten paired seeds, train for ten epochs and report the final epoch. The head trains throughout; LoRA updates begin after three head-only epochs. We resize images directly to $224\times224$, apply ImageNet normalization and use batch size 128. AdamW retains its state across epochs, with head/LoRA learning rates $3\times10^{-4}/10^{-4}$ and weight decay $10^{-4}$. Class-weighted cross-entropy uses label smoothing 0.05, and the backbone remains in evaluation mode.

We define six main pools. Full uses all 37,607 rows. Source-only retains only non-KVASIR training data (10,596 rows); Removal further excludes five other-source rows passing the hash screen against Kvasir-Capsule, leaving 10,591. Random is a fixed 10,596-row sample from Full that matches Source-only's size and nominal update count. Comp (class-count-matched) matches Source-only's class counts; Class-match-R matches Removal's counts at 10,591 rows. Both are drawn from Full, taking KVASIR rows first within each class. Dropping the final incomplete batch leaves 293 minibatches per epoch for Full and 82 for each reduced pool: equal epochs therefore do not imply equal update counts. Paired intervals quantify variation across seeds conditional on fixed lists and pipeline, not patient sampling; each table states whether $\pm$ denotes a sample SD, an SE or the paired-$t$ 95\% half-width $h_{95}$. The records only partly document how lists and hyperparameters were chosen. Appendix~\ref{app:controls} gives the optimizer settings and ordered pool-construction rules.

\paragraph{Evaluation populations.}
We evaluate trained models on public validation (16,132 rows), non-KVASIR validation (4,551) and internal-filtered validation (4,326). The last view removes 225 rows (221 unique basenames) from non-KVASIR validation. A row is removed when its basename (filename without folder) or exact pHash matches a same-source training row; no same-class, dHash or NCC condition is required, and extending the removal to other rows sharing these basenames adds no rows. All ten classes occur in each view. Balanced accuracy (BA) is mean class recall; the organizer score averages class-macro AUC and BA.

\section{What the case reveals}
\label{sec:findings}
\subsection{File identity and video overlap answer different questions}
\label{sec:audit}
Byte-identical files cross the public training--validation boundary in 216 groups: 210 SEE-AI groups involving 215 validation files, and six AIIMS groups involving six files. A filename-only check would miss 45 of them (39 SEE-AI groups and all six AIIMS groups) because their basenames differ across the boundary. No byte-identical files cross this boundary within KVASIR or KID, or across source labels (Figure~\ref{fig:boundary_overview}). Group counts differ from counts of affected rows; Table~\ref{tab:byte_identity} in Appendix~\ref{app:release} gives the complete census by source.

\begin{figure}[htbp]
\centering
\begin{tikzpicture}[x=1pt,y=1pt,font=\fontfamily{phv}\fontsize{8bp}{9bp}\selectfont,
  every node/.style={inner sep=0pt,outer sep=0pt,text=black!85}]
\path[use as bounding box] (0,-6) rectangle (396,224.4);
\definecolor{relBlue}{HTML}{0072B2}
\definecolor{relGray}{HTML}{657581}
\definecolor{relPale}{HTML}{DBE3EA}
\definecolor{relOrange}{HTML}{AD4B20}
\node[anchor=west,font=\fontfamily{phv}\bfseries\fontsize{10bp}{11bp}\selectfont] at (6,218) {(a) File identity};
\node[anchor=west,font=\fontfamily{phv}\bfseries\fontsize{10bp}{11bp}\selectfont] at (218,218) {(b) Video sharing};
\node at (309,194) {41 / 41 prefixes in both splits};

\foreach \value in {0,100,200} {
  \pgfmathsetmacro{\xx}{50+122*(\value/240)}
  \draw[black!7] (\xx,66) -- (\xx,178);
  \draw[black!45] (\xx,66) -- (\xx,63);
  \node[anchor=north] at (\xx,60) {\value};
}
\draw[black!45] (50,66) -- (172,66);
\foreach \name/\yy in {SEE-AI/164,AIIMS/136,KVASIR/108,KID/80}
  \node[anchor=east] at (46,\yy) {\name};
\fill[relGray] (50,158.5) rectangle (66.775,169.5);
\fill[relOrange!23] (66.775,158.5) rectangle (156.75,169.5);
\path[pattern=north east lines,pattern color=relOrange,draw=relOrange,line width=.4pt]
  (66.775,158.5) rectangle (156.75,169.5);
\node[text=white] at (58.3875,164) {33};
\node[fill=relOrange!23,inner sep=1pt] at (111.7625,164) {177};
\node[anchor=west] at (159.75,164) {210};
\fill[relGray] (50,130.5) rectangle (53.05,141.5);
\node[anchor=west] at (56.05,136) {6};
\node[anchor=west] at (52,108) {0};
\node[anchor=west] at (52,80) {0};
\node at (111,44) {Byte-identity groups};
\node[anchor=base west] at (40,21.2) {Label sets:};
\fill[relGray] (88,20) rectangle (96,28);
\node[anchor=base west] at (100,21.2) {Intersect};
\fill[relOrange!23] (144,20) rectangle (152,28);
\path[pattern=north east lines,pattern color=relOrange,draw=relOrange,line width=.4pt]
  (144,20) rectangle (152,28);
\node[anchor=base west] at (156,21.2) {Disjoint};

\foreach \training/\validation [count=\i from 0] in {592/268,686/326,1062/418,478/190,470/202,88/39,143/62,697/293,1188/508,482/226,541/254,221/85,349/150,1464/663,262/108,1435/651,491/205,102/49,511/204,931/377,800/347,357/129,567/209,10/1,912/375,1175/524,984/433,330/119,521/237,911/407,437/184,628/258,1471/648,289/141,830/347,1066/433,562/248,325/170,1467/607,795/335,381/151} {
  \pgfmathsetmacro{\xx}{231+156*\i/41}
  \pgfmathsetmacro{\xr}{231+156*(\i+1)/41-.45}
  \pgfmathsetmacro{\yv}{66+112*(\validation/(\training+\validation))}
  \fill[relPale] (\xx,66) rectangle (\xr,178);
  \fill[relBlue] (\xx,66) rectangle (\xr,\yv);
}
\draw[black!45] (231,66) -- (387,66);
\foreach \value in {0,25,50,75,100} {
  \pgfmathsetmacro{\yy}{66+112*\value/100}
  \draw[black!45] (228,\yy) -- (231,\yy);
  \node[anchor=east] at (225,\yy) {\value};
}
\draw[black!65,dashed,line width=.7pt] (231,99.6) -- (387,99.6);
\foreach \value in {1,10,20,30,41} {
  \pgfmathsetmacro{\xx}{231+156*(\value-.5)/41}
  \draw[black!45] (\xx,66) -- (\xx,63);
  \node[anchor=north] at (\xx,60) {\value};
}
\node[rotate=90] at (202,122) {Rows (\%)};
\node at (309,44) {KVASIR prefix (sorted by ID)};
\node[anchor=base west] at (236,21.2) {Split:};
\fill[relBlue] (262,20) rectangle (270,28);
\node[anchor=base west] at (274,21.2) {Validation};
\fill[relPale] (322,20) rectangle (330,28);
\node[anchor=base west] at (334,21.2) {Training};
\node[align=center,text=black!65] at (198,5) {Prefixes identify recordings;\\patient identity across recordings is unavailable.};
\end{tikzpicture}
\caption{File identity and video sharing across the public training--validation boundary. (a) Byte-identity groups by source and released label relationship. The SEE-AI groups involve 215 validation files; the AIIMS groups involve six. (b) Training and validation row shares within each of the 41 contributing KVASIR video prefixes, sorted by identifier and normalized separately for each prefix. Every prefix occurs in both partitions despite the absence of byte-identical KVASIR crossings. The dashed line marks the announced 30\% validation share.}
\label{fig:boundary_overview}
\end{figure}
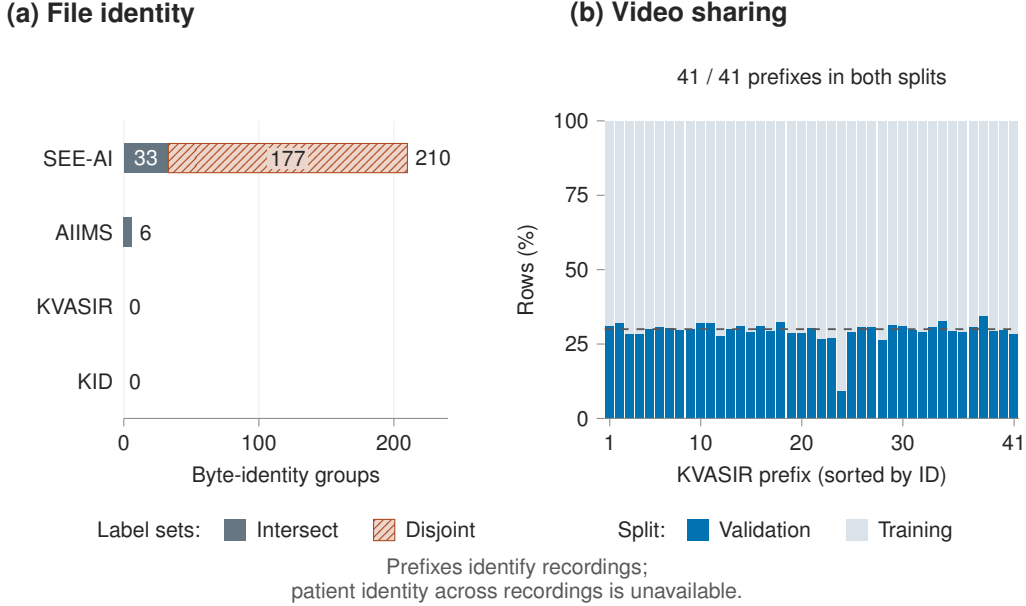

Identical files need not have matching labels: 177 SEE-AI groups have disjoint released label sets, while 33 SEE-AI groups and all six AIIMS groups share a released label. KVASIR illustrates a different distinction: no byte-identical files cross the boundary, yet all 41 contributing video prefixes occur in both partitions. The frame-index gap between each KVASIR validation row and the nearest training frame from the same video has median one and interquartile range 1--1. File identity, label relationships and video sharing thus provide complementary evidence of cross-split dependence.

The separate official-test byte scan finds zero matches against public training, one label-disjoint match against validation and eight within-test groups. Final-test training could use both public partitions, so the absence of training-folder matches does not cover every permitted final-training pool or establish which participants used matching files for training.

\subsection{The counterpart rule changes the measured relationship}
\label{sec:matching}
Exact-pHash matching links 1,381 of 11,581 KVASIR validation rows ($\PhashRate\%$) to same-source training rows in 6,285 pairs. Pairing each validation row with its lexicographically first matching training row, its \emph{counterpart}, yields 540 rows with NCC $\geq0.99$; accepting any matching training row as the counterpart yields 799. The rates among all KVASIR validation rows are $\ChosenNCCRate\%$ and $\AnyNCCRate\%$. With the threshold fixed, changing the counterpart rule changes the question and numerator; the larger count does not establish higher detector recall. On the same 1,381 selected pairs, $256\times256$ bilinear grayscale processing gives 387 NCC-positive pairs, compared with 540 at $96\times96$ bicubic. Resolution, interpolation and the NCC normalization change together (Appendix~\ref{app:detectors}); the 799-row count uses the separate any-counterpart rule.

We compare these similarity scores with same-modality controls. The 1,381 selected exact-pHash pairs have median SSIM 0.880, compared with 0.647 for unflagged adjacent pairs from the same video. Joint video-prefix resampling gives a 95\% interval of $[0.186,0.285]$ for the difference between these SSIM medians. Yet 68 adjacent comparison pairs also reach NCC 0.99, and their duplicate status is unknown. The $\PhashRate\%$ exact-pHash match rate is a census of all KVASIR validation rows, not a sample estimate. Resampling ten of the 41 prefixes per draw gives a $[4.39\%,24.82\%]$ percentile range that describes sensitivity to hypothetically sampling fewer prefixes, not uncertainty in the census or patient membership.

Source-to-reference screening uses the hash screen to compare CV2024 with upstream Kvasir-Capsule, a different boundary. Pixel measurements use a pHash-minimizing counterpart, with ties broken by a fixed enumeration order. Among 38,592 selected KVASIR/reference pairs, 92.31\% reach NCC 0.99; mean DINOv2 cosine similarity is 0.9658, compared with 0.5376 for random pairs. At the same NCC 0.99 cutoff, only 36.17\% pass when the larger of the selected pair's pHash and dHash distances is 3--6. The joint rule misses all tested 90\% center crops in a 50-frame check (Appendix~\ref{app:detectors}).

\begin{figure}[htbp]
\centering\input{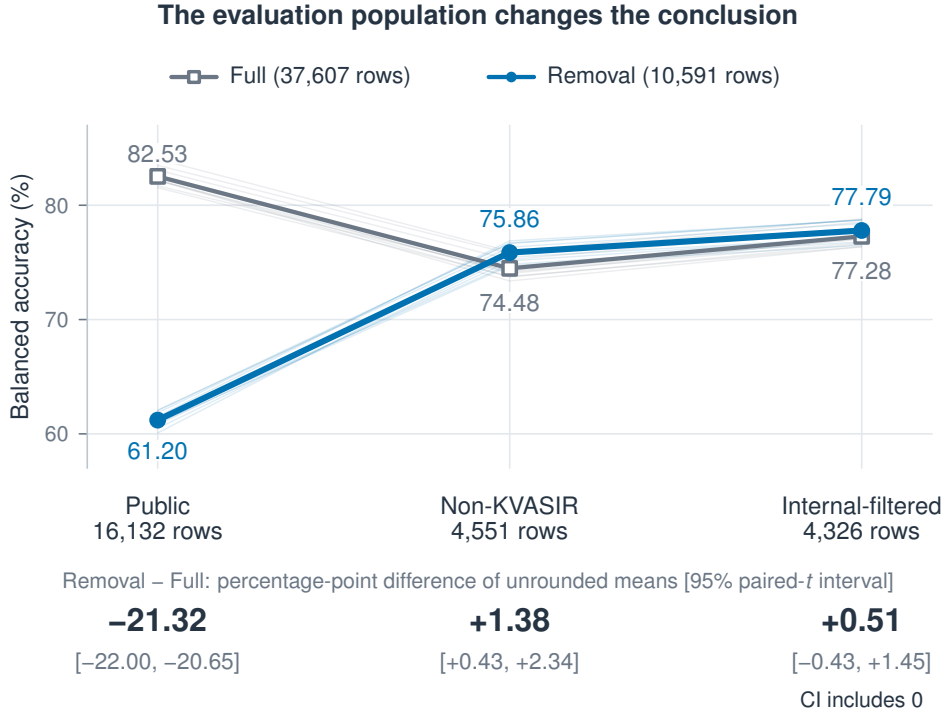}
\caption{A training comparison reverses across evaluation populations (DINOv2-L/14). Full uses 37,607 rows; Removal excludes KVASIR and five other-source rows passing the Kvasir-Capsule hash screen, retaining 10,591. Markers show ten-seed mean BA; faint paths follow each fitted model's scores across views. Full leads on public validation; Removal has higher mean BA on non-KVASIR. Below each view are paired Removal-minus-Full changes in percentage points and pointwise 95\% $t$ intervals conditional on fixed lists. The internal-filtered interval includes zero. Table~\ref{tab:main_evidence} reports all six training pools.}
\label{fig:source_sensitivity}
\end{figure}

\subsection{Matched controls separate sample count from source composition}
\label{sec:counterfactual}
We next ask how removing KVASIR training data changes performance across evaluation populations (Table~\ref{tab:main_evidence}). The central result is a sign reversal: Removal-minus-Full BA is $\RemovalDelta$ on public validation (95\% paired-$t$ interval $\RemovalCI$) but $+0.0138$ ($[+0.0043,+0.0234]$) on non-KVASIR. On internal-filtered validation, the contrast is $\InternalMainDelta$ and its interval includes zero. Figure~\ref{fig:source_sensitivity} follows the same model pairs across these views. The controls below distinguish explanations for this response.

\begin{table}[t]
\centering
\caption{Training and evaluation sensitivity. Public/Non-KV/Filtered: 16,132/4,551/4,326 rows, ten classes each. Non-KV excludes KVASIR; Filtered additionally excludes internal-match candidates (definitions in Section~\ref{sec:design}). BA: mean class recall; AUC: macro one-versus-rest.}
\label{tab:main_evidence}
\definecolor{ClaimTint}{HTML}{E8F0F5}
\setlength{\tabcolsep}{2pt}
\arrayrulecolor{black!65}
\begin{minipage}[t]{0.48\linewidth}\vspace{0pt}
\begin{tabularx}{\linewidth}{@{}>{\raggedright\arraybackslash}Xrrrr@{}}
\multicolumn{5}{@{}l}{\textbf{(a) Training pools (DINOv2, BA)}} \\[2pt]
\rowcolor{black!5} Pool & $n_{\rm train}$ & Public & Non-KV & Filt. \\
\midrule[0.4pt]
Full & 37,607 & \cellcolor{ClaimTint}$.825$ & \cellcolor{ClaimTint}$.745$ & \cellcolor{ClaimTint}$.773$ \\
Random & 10,596 & $.779$ & $.695$ & $.707$ \\
Comp & 10,596 & $.734$ & $.388$ & $.401$ \\
Class-match-R & 10,591 & $.738$ & $.391$ & $.406$ \\
Removal & 10,591 & \cellcolor{ClaimTint}$.612$ & \cellcolor{ClaimTint}$.759$ & \cellcolor{ClaimTint}$.778$ \\
Source-only & 10,596 & $.611$ & $.757$ & $.776$ \\
\bottomrule[0.4pt]
\end{tabularx}
\par\vspace{12pt}
\begin{tabularx}{\linewidth}{@{}>{\raggedright\arraybackslash}Xrr@{}}
\multicolumn{3}{@{}l}{\textbf{(b) Training contrasts ($\Delta$BA)}} \\[2pt]
\rowcolor{black!5} Contrast & Public & Non-KV \\
\midrule[0.4pt]
Rem.$-$Full & $-.2132\pm.0068$ & $+.0138\pm.0096$ \\
S$-$R & \cellcolor{ClaimTint}$-.1682\pm.0072$ & \cellcolor{ClaimTint}$+.0624\pm.0108$ \\
A$-$Comp & $-.0690\pm.0062$ & $+.0978\pm.0079$ \\
R50: S$-$R & $-.2002\pm.0023$ & $+.0878\pm.0042$ \\
CNX: S$-$R & $-.2096\pm.0041$ & $+.0955\pm.0035$ \\
\bottomrule[0.4pt]
\end{tabularx}
\end{minipage}\hfill\begin{minipage}[t]{0.48\linewidth}\vspace{0pt}
\begin{tabularx}{\linewidth}{@{}>{\raggedright\arraybackslash}Xrr@{}}
\multicolumn{3}{@{}l}{\textbf{(c) Official test ($\mu\pm\mathrm{SE}$)}} \\[2pt]
\rowcolor{black!5} Pool & BA & AUC \\
\midrule[0.4pt]
Full & $.2804\pm.0046$ & $.7177\pm.0015$ \\
Random & $.2786\pm.0038$ & $.7352\pm.0036$ \\
Source-only & $.2688\pm.0034$ & $.7533\pm.0013$ \\
\bottomrule[0.4pt]
\end{tabularx}
\par\vspace{6pt}
\begin{tabularx}{\linewidth}{@{}>{\raggedright\arraybackslash}Xrr@{}}
\multicolumn{3}{@{}l}{\textbf{(d) Fixed predictions: $\Delta$combined score}} \\[2pt]
\rowcolor{black!5} Cohort / aggregation & Non-KV & Filtered \\
\midrule[0.4pt]
Pooled 24 & $-.0503$ & $-.0393$ \\
Equal-source 24 & $-.0199$ & $-.0163$ \\
Pooled 19 & $-.0624$ & $-.0488$ \\
Equal-source 19 & $-.0220$ & $-.0177$ \\
\bottomrule[0.4pt]
\end{tabularx}
\par\vspace{6pt}
\begin{tabularx}{\linewidth}{@{}>{\raggedright\arraybackslash}Xrr@{}}
\multicolumn{3}{@{}l}{\textbf{(e) Local$-$distant, exposed$-$unflagged}} \\[2pt]
\rowcolor{black!5} Endpoint & $\Delta\pm h_{95}$ & Signs \\
\midrule[0.4pt]
True-class prob. & \cellcolor{ClaimTint}$-.00355\pm.00131$ & 10/0/0 \\
BA (6 classes) & \cellcolor{ClaimTint}$-.00194\pm.00291$ & 4/6/0 \\
Accuracy & \cellcolor{ClaimTint}$-.00079\pm.00084$ & 4/6/0 \\
\bottomrule[0.4pt]
\end{tabularx}
\end{minipage}
\par\vspace{6pt}
\begin{minipage}{\linewidth} \textbf{Filtered-view contrast:} Removal$-$Full $\Delta$BA = \colorbox{ClaimTint}{$+.0051\pm.0094$}.\par\smallskip Blue marks key comparisons, not significance. Absolute units; ten seeds in (a--c,e). (b,e) and footer report mean $\pm$ paired-$t$ 95\% half-width $h_{95}$ (not SE), conditional on fixed lists. First three rows in (b): DINOv2-L/14; R50/CNX: frozen ResNet-50/ConvNeXt-Tiny. Rem.: Removal; S/R: Source-only/Random (the R of Class-match-R denotes Removal); A: 66 Ulcer KVASIR-to-AIIMS replacements. (c) Separate experiment evaluated on the 4,385-row, ten-class official test; mean $\pm$ SE. (d) Combined-score change from the Public view under the same aggregation: pooled $(\mathrm{BA}+\mathrm{AUC})/2$, or equal-source averages with represented-class support. Descriptive; entries may be related: 24 score-consistent entries (excluding only Team G) are primary; 19 are a fixed subset sensitivity (Appendix~\ref{app:public_predictions}). (e) Exploratory exposed-minus-unflagged, local-minus-distant contrast on 588 exposed and 588 matched-unflagged rows; signs: negative/zero/positive seeds (Section~\ref{sec:intervention}). \end{minipage}
\end{table}

\paragraph{Matching training size and updates.}
Random and Source-only both contain 10,596 rows and have equal nominal update counts. Their public BA values are 0.779 and 0.611, respectively. The Source-only-minus-Random contrast is $-0.1682$ on public validation and $+0.0624$ on non-KVASIR (Table~\ref{tab:main_evidence}(b)), so sample count alone cannot explain the public loss or its reversal. Removal produces a similar public loss to Source-only: the five additional exclusions change BA by $\FiveRowDelta$ (95\% interval $\FiveRowCI$).

\paragraph{Class counts and within-class source replacement.}
Comp and Class-match-R preserve the corresponding reduced pool's class counts, but 63\% of their rows are KVASIR and the dominant source differs by class, so source appearance can predict class (Appendix~\ref{app:composition}). They reach public BA 0.734 and 0.738 but only 0.388 and 0.391 on non-KVASIR, compared with 0.745 for Full. Pool A replaces Comp's 66 KVASIR Ulcer images with 66 AIIMS Ulcer images and retains every non-Ulcer row. Public BA falls by 0.069 (95\% interval $[-0.0752,-0.0628]$), whereas BA rises by 0.098 ($[+0.0898,+0.1057]$) on non-KVASIR, whose Ulcer rows are all AIIMS. This replacement changes the score even when sample and class counts are fixed, so composition is essential to interpreting removal effects.

\paragraph{Comparisons across model families, sources and classes.}
Frozen ResNet-50~\citep{he2016resnet} and ConvNeXt-Tiny~\citep{liu2022convnext} with nonlinear heads show the same reversal: Source-only-minus-Random is $-0.200$ and $-0.210$ on public validation, respectively, and positive on non-KVASIR. For the primary pipeline, source-specific Removal-minus-Full BA changes are $-0.473$ (KVASIR), $+0.013$ (SEE-AI), $-0.044$ (KID) and $+0.021$ (AIIMS). These source-specific averages cover 9, 8, 5 and 2 classes, respectively, and are not additive components of pooled BA. Public-validation Ulcer recall falls from 0.976 to 0.096, and recall losses vary across classes. The matched-size reversal occurs in all three tested model families, alongside source- and class-specific variation.

Auxiliary analyses are consistent with source dependence but do not separate domain familiarity from image-level exposure. Removing only rows whose source-matched NCC reaches 0.99 or 0.90 retains 12,697 and 10,613 training rows (2,101 and 17 KVASIR rows), giving public BA 0.744 and 0.654. Broader removal changes both content and size. Permuting KVASIR labels while preserving their histogram gives BA 0.504; this tests label sensitivity, not instance memorization. Removing SEE-AI lowers SEE-AI validation BA from 0.697 to 0.239. These auxiliary comparisons use different hardware and lack matched-size controls (Appendix~\ref{app:sensitivity_controls}).

\paragraph{Metric and endpoint sensitivity.}
Full and Removal have mean public AUC $\TrainingFullAUC$ and $\TrainingRemovalAUC$, and combined scores $\TrainingFullCombined$ and $\TrainingRemovalCombined$ (change $\TrainingCombinedDelta$). Using each run's highest public-validation BA instead of its epoch-ten value gives $\TrainingFullBestBA$ and $\TrainingRemovalBestBA$. The contrast is $\TrainingBestDelta$, close to the final-epoch contrast of $\TrainingFinalDelta$. The public loss thus appears in both score components and under both epoch choices. Selecting an epoch using evaluation outcomes makes that comparison descriptive; the final epoch remains the common reporting point. On non-KVASIR, Removal-minus-Full is positive for the combined score ($+0.0062$) but negative for AUC ($-0.0015$) and accuracy ($-0.0138$), with all three 95\% paired-$t$ intervals excluding zero; Source-only-minus-Random is positive on all three.

\subsection{Fixed-prediction diagnostics expose metric and rank sensitivity}
\label{sec:rescore}
The preceding experiments change training; re-scoring the participants' submitted predictions holds them fixed and gives an analogue of Eq.~\ref{eq:view}. Of the 27 released workbooks, 25 are scoreable; 24 of these agree with their reported full-list scores and form our primary, score-consistent cohort. Score agreement establishes neither that public validation was held out from model fitting and selection nor that the same checkpoint generated both the public-validation and official-test predictions. The primary cohort's mean organizer-score change on non-KVASIR is $\ScoreConsistentPooledNonKV$, with 19 entries changing rank. A 19-entry subset omitting Teams U--Y gives $\TrainedMSeven$, with 14 entries changing rank. Its original selection rationale is unverified, so it serves only as a sensitivity analysis (Appendix~\ref{app:public_predictions}).

For the same 24 entries, equal-source averaging gives $\ScoreConsistentEqualNonKV$, compared with the pooled change of $\ScoreConsistentPooledNonKV$ (Table~\ref{tab:main_evidence}(d)). Within each source, BA averages recall over represented classes and AUC averages the defined one-versus-rest comparisons. We then average the combined scores equally across sources: four in the public view and three in the non-KVASIR and filtered views. The contrast therefore depends on both source weighting and class support. Appendix~\ref{app:public_predictions} specifies cohort exclusions and resampling sensitivities.

Spearman correlation with official-test combined scores falls from $\ScoreConsistentPublicRho$ on public validation to $\ScoreConsistentNonKVRho$ on non-KVASIR, and it also falls in each sensitivity cohort (Table~\ref{tab:rank_alignment}). A separate DINOv2 official-test study shows that the two score components can respond differently (Table~\ref{tab:main_evidence}(c)). Relative to Full, Source-only training raises mean AUC by 0.0356 (95\% paired-$t$ interval $[+0.0309,+0.0403]$; all ten seeds), whereas the interval for its BA difference, $-0.0117$ ($[-0.0259,+0.0026]$), includes zero. With the same training recipe, the separate official-test fits reach BA of only 0.27--0.28, against 0.61--0.83 for the primary fits on public validation. Reporting diagnostic views alongside the official result makes this dependence on population and metric explicit~\citep{maierhein2018ranking}.

\subsection{Local versus distant removal: an exploratory probability contrast}
\label{sec:intervention}
This experiment asks whether removing the training neighbors of exposed validation frames (frames with an exact-pHash training counterpart at NCC $\geq0.99$) lowers confidence on those frames more than removing equally many distant training rows does, relative to matched unflagged frames. We compare local and distant removals within video/class strata while retaining KVASIR. All arms first remove the 1,100 distinct training rows linked to the initial 799 exposed rows using this matching rule. Strata need enough distant rows to match local removal; one fails, excluding 211 exposed rows and leaving 588 exposed rows $E$ and 588 matched-unflagged rows $U$. We sampled $U$ without replacement to match video/class counts before eligibility filtering. Unflagged rows have no exact-pHash counterpart, which does not verify absence of exposure. Each group contains 547 Normal rows and 41 rows from five abnormal classes.

\begin{table}[htbp]
\centering
\setlength{\tabcolsep}{2pt}
\arrayrulecolor{black!65}
\caption{Local and distant removal: absolute outcomes and true-class probability contrasts. E: exposed; U: matched-unflagged (588 rows each). For Distant, we average the three fixed control arms within each seed, then across the ten seeds. BA averages recall over the same six classes.}
\label{tab:local_endpoints}
\textbf{(a) Absolute endpoint means (\%)}\par\smallskip
\begin{tabular}{@{}lrrrr@{}}
\toprule[0.4pt]
\rowcolor{TableHeader} Endpoint & E local & E distant & U local & U distant \\
\midrule[0.4pt]
True-class probability & $90.654$ & $90.975$ & $89.670$ & $89.636$ \\
Accuracy & $99.949$ & $99.977$ & $99.609$ & $99.558$ \\
Fixed six-class BA & $99.667$ & $99.852$ & $94.947$ & $94.938$ \\
\bottomrule[0.4pt]
\end{tabular}
\par\medskip
\textbf{(b) Probability: local minus distant (percentage points)}\par\smallskip
\begin{tabular}{@{}lrrr@{}}
\toprule[0.4pt]
\rowcolor{TableHeader} Component & Mean & Paired-$t$ 95\% interval & Signs ($- / 0 / +$) \\
\midrule[0.4pt]
Exposed (E) & $-0.321$ & $[-0.538, -0.104]$ & 8/0/2 \\
Unflagged (U) & $+0.034$ & $[-0.141, +0.209]$ & 4/0/6 \\
E$-$U & $-0.355$ & $[-0.486, -0.224]$ & 10/0/0 \\
\bottomrule[0.4pt]
\end{tabular}
\par\smallskip
\begin{minipage}{\linewidth} Signs count negative/zero/positive seed contrasts (zero tolerance $10^{-12}$). Intervals use ten paired seed blocks after averaging distant arms, not 30 independent controls. Both strategies remove training rows; no unremoved baseline is compared. Design, seed set, controls and endpoint were selected with access to results. Intervals are exploratory, conditional on those choices and unadjusted for selection; they do not establish patient-level generalization. \end{minipage}
\end{table}

Within eligible strata, each arm removes 876 additional rows: local neighbors in one arm, or distant rows matched by video/class in each of three fixed control arms. Eligibility filtering leaves the common removal unchanged; all arms retain 35,631 rows. Local removal selects each counterpart's immediate predecessor and successor in its video/class list, sorted by frame index. Distant controls match the resulting run lengths, exclude the local set, stay at least 50 frame indices from exposed rows and apply additional hash-match exclusions (Appendix~\ref{app:intervention}). Adjacent entries in a list need not be consecutive source frames.

For endpoint $g$, local arm $L$, distant arms $C_j$ and paired seed $s$, we estimate
\begin{equation}
D_s(g)=\left[g_{s,L}(E)-\frac{1}{3}\sum_{j=1}^{3}g_{s,C_j}(E)\right]
-\left[g_{s,L}(U)-\frac{1}{3}\sum_{j=1}^{3}g_{s,C_j}(U)\right].
\label{eq:did}
\end{equation}
This analysis uses epoch ten and ten paired seeds; its design and endpoint were selected with access to evaluation outcomes (Appendix~\ref{app:intervention}). True-class probability, negative log-likelihood (NLL), probability margin and accuracy are averaged over rows; BA averages recall over the six represented classes. Probability margin is true-class probability minus the largest alternative-class probability.

Table~\ref{tab:local_endpoints} reports absolute outcomes and the two components of Eq.~\ref{eq:did}. Local-minus-distant probability changes are $\LocalExposedProbabilityPP$ percentage points for $E$ and $\LocalUnflaggedProbabilityPP$ for $U$; the interval for $U$ spans zero. Their paired difference is $-\ProbDropPP$ (95\% paired-$t$ interval $[\ProbCILoPP,\ProbCIHiPP]$). This difference is negative in all ten seeds, although the exposed component is negative in eight. The NLL contrast is $+0.00495$ (95\% paired-$t$ interval $[+0.00329,+0.00660]$), and the probability-margin contrast is $-0.00472$ ($[-0.00647,-0.00297]$). NLL uses natural-log units; the margin contrast uses absolute probability units. BA and accuracy intervals span zero (Table~\ref{tab:main_evidence}(e)); the four group--strategy accuracy means range from $\LocalAccuracyMinPercent$\% to $\LocalAccuracyMaxPercent$\%.

The rows of $E$ and $U$ come from 32 video prefixes. Omitting each prefix in turn leaves the mean probability contrast negative, ranging from $\LocalProbabilityLOPOMinPP$ to $\LocalProbabilityLOPOMaxPP$ percentage points; this range describes omission-specific means, not confidence limits. The three largest-magnitude signed prefix contributions account for $\LocalTopThreePrefixSharePercent\%$ of the net contrast. Each minority class occurs in only one prefix, and only 914 of 10,000 prefix-bootstrap draws retain all six classes; because averaging over present classes changes the BA target, we report no fixed six-class BA interval under prefix resampling (Appendix~\ref{app:intervention}).

\section{ReuScope: audit design and interpretation}
\label{sec:protocol}
\label{sec:discussion}
Table~\ref{tab:audit_records} links the CV2024 observations to supported conclusions. ReuScope organizes this audit into four steps, starting from released lists, labels and grouping metadata. Access to images, predictions and training determines which analyses are possible.

\begin{table}[htbp]
\centering
\setlength{\tabcolsep}{2pt}
\arrayrulecolor{black!65}
\caption{Selected conclusions from the CV2024 audit. The counterpart comparison counts KVASIR validation rows with exact-pHash matches in KVASIR training. Missing information differs from a measured zero; populations and uncertainty are defined with each analysis.}
\label{tab:audit_records}
\begin{tabular}{@{}>{\raggedright\arraybackslash}p{0.22\linewidth}>{\raggedright\arraybackslash}p{0.35\linewidth}>{\raggedright\arraybackslash}p{0.37\linewidth}@{}}
\toprule[0.4pt]
\rowcolor{TableHeader} Question & Observation & Supported conclusion \\
\midrule[0.4pt]
Identical content? & SEE-AI: 210 groups, 215 validation files; AIIMS: 6 groups, 6 files. & Released bytes cross the public boundary. Label relationships are assessed separately. \\
\addlinespace
Shared videos? & KVASIR: 0 byte crossings; all 41 contributing prefixes shared. & The split separates files while retaining source-video overlap. \\
\addlinespace
Similar to which counterpart? & NCC $\geq0.99$: 540 chosen-counterpart vs. 799 any-match rows. & Different matching rules count different relationships. \\
\addlinespace
Training sensitivity? & Removal-minus-Full BA: $-0.2132$ public, $+0.0138$ non-KVASIR. & The training contrast depends on the evaluation population. \\
\addlinespace
Patient-level linkage? & Shared recordings observed; cross-recording patient IDs unavailable. & Unique-patient overlap cannot be quantified from public metadata. \\
\bottomrule[0.4pt]
\end{tabular}
\end{table}

\FloatBarrier

\paragraph{1. Define the boundary and evidence unit.}
Specify the release, query/reference populations, the information permitted for model development and the intended generalization claim. Preserve row identities, multiplicities and labels. State what each grouping key, such as a video prefix, denotes and which identity levels, such as patient, are unavailable.

\paragraph{2. Measure each data relationship separately.}
With images, confirm positive cryptographic-hash groups byte for byte and assess label relationships separately. Use documented identifiers to measure group overlap. For perceptual similarity, specify preprocessing, the screening predicate, counterpart selection, search coverage, same-modality comparison pairs and operating limits. Report pairs, identity groups and affected rows with their own denominators.

\paragraph{3. Separate training and evaluation changes.}
Aligned prediction records support comparisons across evaluation views with predictions held fixed. With training access, compare matched pools across these views. Report the original public view alongside source-removed and stricter internal-filtered diagnostics where definable; these do not replace the official leaderboard. State sample count, class/source support, update budget, endpoint and seed pairing. Match controls to alternative explanations: Source-only versus Random fixes sample size and the nominal update budget; the within-class replacement fixes size and class counts (Section~\ref{sec:counterfactual}). If a required class is absent from an evaluation set or resample, mark the fixed-class statistic undefined; averaging over present classes defines a different target. Intervals across seeds on fixed lists and group resampling address different sources of uncertainty. Record outcome-informed choices and targets that cannot be estimated.

\paragraph{4. Report conclusions with their scope and uncertainty.}
For each analysis, record its boundary, predicate or contrast, counting unit, denominator, coverage, class support, estimate, uncertainty, selection scope, supported conclusion and unresolved target. Link the record to input lists and numerical evidence. Mark its status as completed, partial, not assessed or not applicable; a measured zero differs from missing evidence. When adapting grouping definitions, screens and training/evaluation interfaces, preserve these fields and the distinction between data relationships and score responses.

\paragraph{A worked audit record.}
For the any-counterpart finding, the boundary is KVASIR public training versus its 11,581 validation rows. The exact-pHash/NCC $\geq0.99$ rule flags 799 validation rows; choosing one lexicographic counterpart flags 540. The record establishes candidate correspondence under each rule. We record changes in model performance separately (Eq.~\ref{eq:train}); patient linkage across recordings remains unknown.

\paragraph{Reproduction and responsible reporting.}
Code and data for the main analyses, including exact input lists, saved predictions and numerical records, are publicly released; Appendix~\ref{app:reproducibility} gives the repository, maps results to code and data, and lists the components that are not released. Where feasible and compatible with review anonymity, share metadata-only findings with the audited party before submission, invite factual corrections and report when this occurred and what was clarified. Report organizer clarifications separately from independently verified observations. Misinterpreting release anomalies as individual misconduct could cause reputational harm to organizers or participants; report release properties without inferring clinical label correctness or individual responsibility. We contacted the organizers only after the initial submission; Appendix~\ref{app:release} documents the exchange.

\paragraph{Applying components to other releases.}
The ChestX-ray14 lists contain 86,524 train/validation and 25,596 test filenames, with 28,008 and 2,797 patient prefixes. Neither filenames nor prefixes overlap~\citep{wang2017chestxray}. This checks the complete identifier lists, not pixel content. For image screening, we use 25,331 ISIC 2019 training images~\citep{isic2019data} and 10,662 labeled HyperKvasir images~\citep{borgli2020hyperkvasir}. Across strata defined by the ISIC lesion-ID prefix, the sum rule yields three candidates, whereas the joint rule yields 67. Of the 67, two pass NCC $\geq0.99$ at both tested resolutions, and one is also byte-identical. Each of the two pairs links an image with an empty lesion ID to one with an MSK-prefixed ID, so these strata are metadata groups rather than verified institutions. Comparing HyperKvasir with CV2024 KVASIR yields no candidate under the sum rule and one under the joint rule; that pair fails NCC 0.99. Appendix~\ref{app:external} gives the pair breakdown. These checks apply the identifier and image-screening components to other releases; they do not test the full sequence of training and evaluation comparisons.

To apply ReuScope elsewhere, adapt the source/group mappings and screen to the new release while retaining explicit boundaries, controls and result fields; the released code encodes CV2024's sources, classes and Ulcer-replacement counts, which must be replaced. Prevalence, thresholds and effect sizes remain specific to each application.

\section{Conclusion}
\label{sec:conclusion}
CV2024 shows why benchmark audits must connect data relationships to defined comparisons. Filename, byte and video checks reveal different dependencies; a source-removal BA contrast reverses across evaluation populations; and the exploratory local-versus-distant comparison yields a negative difference-in-differences in true-class probability. These findings inform ReuScope's separation of training and evaluation changes and its reporting of each estimate's boundary, population and uncertainty.

\paragraph{Limitations.} The audit covers one benchmark and one primary training pipeline; the frozen-CNN comparisons and other-release checks each test only part of the protocol. Shared recordings establish overlap across the public split, but missing cross-recording identifiers limit patient-level quantification. Intervals describe seed variation on fixed lists, not patient sampling. The local comparison was designed with outcome access, and the similarity screens are not calibrated duplicate detectors.

\label{main:content_end}

\begin{ack}
This work was supported by Artificial intelligence industrial convergence cluster development project funded by the Ministry of Science and ICT (MSIT, Korea) \& Gwangju Metropolitan City and the IITP AI Semiconductor Program and the ANCHOR Program, funded by the Korean government and the Seoul Metropolitan Government (IITP-2023-RS-2023-00256081; 2026-ANCHOR-01-110).

\paragraph{Competing interests.} The authors declare no competing interests.
\end{ack}

\bibliographystyle{unsrtnat}
\bibliography{references}
\clearpage
\appendix
\addtocontents{toc}{\protect\setcounter{tocdepth}{2}}
\section*{Appendix contents}
Appendix~\ref{app:detectors} covers data relationships and grouping sensitivity; Appendix~\ref{app:rescore} covers fixed-prediction diagnostics and official-test evaluation. Appendices~\ref{app:controls}--\ref{app:intervention} describe training comparisons and local removal. Appendix~\ref{app:kvasir_protocols} treats the upstream Kvasir-Capsule experiments as a separate evaluation setting from CV2024. Appendix~\ref{app:external} presents external component checks. Appendix~\ref{app:reproducibility} describes code and numerical-data availability and an audit record.

\makeatletter
\@starttoc{toc}
\makeatother
\clearpage

\section{Data, identity and similarity measurements}
\label{app:detectors}
This section distinguishes two boundaries: CV2024 public training versus validation, and CV2024 images versus the upstream Kvasir-Capsule reference. For each boundary, we specify the populations, counterpart rules and counting units below. We then examine sensitivity to video-prefix grouping.
\subsection{Released populations, identity and grouping}
\label{app:release}
The public CV2024 workbooks define 37,607 training and 16,132 validation rows. Source labels come from the organizer workbooks' \texttt{Sheet1} column \texttt{Dataset}, keyed by full \texttt{image\_path} in workbook order. All 53,739 source labels are recognized and agree with the source directories. Aggregate source counts reproduce the challenge paper's Table I: training/validation counts are KVASIR 27,011/11,581, SEE-AI 9,996/4,289, KID 376/165 and AIIMS 224/97. The hash screen below does not redefine these source labels. Joins preserve full paths, folder labels and workbook order. We group files by SHA-256 digest and compare files within every positive cross-partition group byte for byte. Table~\ref{tab:byte_identity} reports this census separately as identity groups and affected validation rows. Shared folder labels establish that released annotations agree, not that they are clinically correct.

We shared metadata-only findings, including the official-test scan, with the CV2024 organizers. The lead organizer replied that the 70--30 split was made within abnormality-wise source folders without video or case grouping, that duplicate control removed one of two visually similar sequential frames, without cryptographic checks, and that the organizers hold no patient- or examination-level labels for the Kvasir-Capsule frames; she agreed that we report these points and has sent no corrections as of October 2026. The file-identity counts reported here come from our own byte comparisons.

\begin{table}[htbp]
\centering
\setlength{\tabcolsep}{4pt}
\arrayrulecolor{black!65}

\caption{Byte-identical public train--validation crossings. Rates are proportions of affected validation files; a group can contain several files. The final columns count groups whose training and validation label sets intersect or are disjoint. Renamed groups have different training and validation basenames, so a filename-only check misses them. Label-disjoint groups are release/label anomalies; label intersection does not establish clinical label truth.}
\label{tab:byte_identity}

\begin{tabular}{lrrrrrrr}
\toprule[0.4pt]
\rowcolor{TableHeader} Source & \shortstack{Validation\\rows} & Groups & \shortstack{Validation\\files} & Rate & \shortstack{Labels\\intersect} & \shortstack{Labels\\disjoint} & \shortstack{Renamed\\groups} \\
\midrule[0.4pt]
KVASIR & 11,581 & \cellcolor{ClaimTint} 0 & \cellcolor{ClaimTint} 0 & 0.00\% & 0 & 0 & 0 \\
SEE-AI & 4,289 & \cellcolor{ClaimTint} 210 & \cellcolor{ClaimTint} 215 & 5.01\% & 33 & 177 & 39 \\
KID & 165 & 0 & 0 & 0.00\% & 0 & 0 & 0 \\
AIIMS & 97 & \cellcolor{ClaimTint} 6 & \cellcolor{ClaimTint} 6 & 6.19\% & 6 & 0 & 6 \\
\bottomrule[0.4pt]
\end{tabular}

\end{table}

Video prefixes come from attributed source filenames. All 41 contributing KVASIR prefixes occur in both public partitions; this establishes shared examination recordings. The public identifiers do not determine the number of distinct patients or additional overlap across different recordings. Of the 43 labeled reference videos, 21 have more than 90\% attributed-row coverage, defined as attributed CV2024 rows divided by the reference inventory for that prefix. Because the numerator counts rows, this need not be the fraction of unique reference images. After mapping source categories to CV2024 labels (for example, \texttt{blood\_fresh} to Bleeding), labels agree for the 38,584 unambiguously mapped rows; eight filenames that occur under both Erosion and Pylorus in the reference are labeled Erosion in CV2024.

\subsection{Source-to-reference screens and pair selection}
Source-to-reference screening compares CV2024 queries with the upstream Kvasir-Capsule reference corpus, a different boundary from public training--validation overlap. Let $d_p$ and $d_d$ be Hamming distances between 64-bit pHash and dHash descriptors computed with ImageHash~\citep{imagehash}. For query $x$ and reference corpus $R$, the hash screen requires
\[
 \min_{r\in R}d_p(x,r)\leq6
 \quad\text{and}\quad
 \min_{r\in R}d_d(x,r)\leq6.
\]
The joint rule instead requires a single $r$ satisfying both inequalities. All KVASIR queries pass both rules at radius six in this analysis, but they differ at smaller radii and for other-source candidates. Table~\ref{tab:phash_audit} uses independent minima. PDQ provides a complementary representation~\citep{pdq2019}; its distance-50 cutoff is specific to this study and has no capsule-image sensitivity or specificity calibration.

\begin{table}[htbp]
\centering
\setlength{\tabcolsep}{2pt}
\arrayrulecolor{black!65}

\caption{Query-level matching to Kvasir-Capsule. pHash and dHash are minimized independently over reference images. Both minima can be zero without a shared reference match or byte identity. PDQ is a separate perceptual hash; its cutoffs are study-specific. Distance 90 illustrates liberal candidate selection; it is not an official default.}
\label{tab:phash_audit}

\begin{tabular}{lrrrrrr}
\toprule[0.4pt]
\rowcolor{TableHeader} Source & $n$ & \shortstack{Both minima\\zero} & \shortstack{Each min\\$\leq2$} & \shortstack{Each min\\$\leq6$} & PDQ $\leq50$ & PDQ $\leq90$ \\
\midrule[0.4pt]
KVASIR & 38,592 & 23,444 & 38,259 & 38,592 & 38,357 & 38,592 \\
SEE-AI & 14,285 & 0 & 0 & 3 & 0 & 451 \\
KID & 541 & 0 & 0 & 1 & 0 & 39 \\
AIIMS & 321 & 0 & 0 & 1 & 0 & 320 \\
\bottomrule[0.4pt]
\end{tabular}

\end{table}

For pair-level comparisons, we use the pHash-minimizing counterpart recorded in the retained measurements. Equal pHash distances are resolved by first occurrence in the preserved reference-cache order, rather than by a lexicographic path rule. The cache follows filesystem enumeration order; reproducing ties requires that retained ordering, not a newly sorted reference list. Both distances are computed for that same counterpart. The selected counterpart can therefore have dHash distance greater than six even when the query passes the hash screen through another reference. Separate zero minima need not share a counterpart; even a common zero-distance counterpart need not be byte-identical.

For centered grayscale intensity vectors $a,b$, normalized cross-correlation is
\[
 \operatorname{NCC}(a,b)=
 \frac{a^\top b}{\lVert a\rVert_2\lVert b\rVert_2+10^{-12}}.
\]
We use float32 arithmetic after direct $96\times96$ bicubic resizing, without registration or cropping. Pixel scores are computed for the specified counterpart; independently minimized hash distances need not describe that same pair.

Across the 38,592 selected KVASIR/reference pairs, 35,625 (92.31\%) reach NCC $\geq0.99$. Table~\ref{tab:ncc_bands} groups those scored pairs by their own $\max(d_p,d_d)$. Its distance-3--6 band contains 2,372 pairs, of which 858 (36.17\%) reach 0.99. A different grouping---queries whose independently minimized distances have a maximum in the range 3--6---contains 333 queries, with 313 (93.99\%) reaching the NCC threshold with their selected counterpart. These strata describe different populations; neither estimates detector precision. In particular, the high rate in the query-minimum stratum does not describe all pairs whose own distances lie in that band.

\begin{table}[htbp]
\centering
\setlength{\tabcolsep}{2pt}
\arrayrulecolor{black!65}

\caption{Normalized cross-correlation (NCC) by selected-pair hash distance. Both hash distances refer to the pair scored by NCC. The final columns give the fraction of pairs reaching each NCC threshold. Query-level minima can correspond to different reference images and do not define these bands.}
\label{tab:ncc_bands}

\begin{tabular}{lrrrrrr}
\toprule[0.4pt]
\rowcolor{TableHeader} $\max(d_p,d_d)$ & $n$ & Mean NCC & Median NCC & $\geq.99$ & $\geq.95$ & $\geq.90$ \\
\midrule[0.4pt]
0 & 20,448 & 0.9994 & 0.9996 & 99.39\% & 99.99\% & 100.00\% \\
1--2 & 15,473 & 0.9979 & 0.9995 & 93.23\% & 99.76\% & 99.96\% \\
3--6 & 2,372 & 0.9822 & 0.9862 & 36.17\% & 96.21\% & 99.07\% \\
>6 & 299 & 0.9690 & 0.9754 & 6.35\% & 86.96\% & 99.00\% \\
\bottomrule[0.4pt]
\end{tabular}

\end{table}

Feature comparisons use the retained DINOv2-L/14 \texttt{dinov2\_vitl14\_pretrain.pth} checkpoint (LVD-142M).\footnote{Checkpoint source: \url{https://dl.fbaipublicfiles.com/dinov2/dinov2_vitl14/dinov2_vitl14_pretrain.pth}.} They provide a complementary representation of the same upstream-source relationship. We directly resize to $224\times224$, apply ImageNet normalization, extract final-layer-normalized CLS features, normalize their length and compute cosine similarity. The 38,592 KVASIR/reference pairs have mean cosine 0.9658; 1,000 randomly drawn comparison pairs, excluding each query's designated reference, have mean 0.5376. The five other-source screen-positive candidates and random-pair similarity summaries are reported in Table~\ref{tab:features}. These measurements characterize resemblance under a specified representation, not duplicate ground truth or independence from pretraining exposure.

\begin{table}[htbp]
\centering
\setlength{\tabcolsep}{2pt}
\arrayrulecolor{black!65}

\caption{DINOv2 feature cosine similarity for CV2024 queries paired with Kvasir-Capsule references. All similarities use the same feature pipeline; the pHash rows partition the KVASIR pair inventory by hash distance. Threshold percentages describe feature-space similarity, not duplicate labels.}
\label{tab:features}

\begin{tabular}{lrrr}
\toprule[0.4pt]
\rowcolor{TableHeader} Pool & $n$ & Mean cosine & \% cosine $\geq.90$ \\
\midrule[0.4pt]
KVASIR all & 38,592 & 0.9658 & 98.32 \\
pHash 0 & 32,854 & 0.9662 & 98.49 \\
pHash 1--2 & 5,673 & 0.9635 & 97.43 \\
pHash 3--4 & 65 & 0.9509 & 92.31 \\
SEE-AI & 3 & 0.7104 & 0.00 \\
KID & 1 & 0.3928 & 0.00 \\
AIIMS & 1 & 0.4075 & 0.00 \\
Random comparison & 1,000 & 0.5376 & 0.30 \\
\bottomrule[0.4pt]
\end{tabular}

\end{table}

\FloatBarrier

The joint rule can miss transformed copies (Table~\ref{tab:detection_limits}). In a test on 50 real frames with synthetic transforms, no pair passes the radius-six condition after any tested center crop or a $10^\circ$ rotation, only 5 of 50 pass after a $5^\circ$ rotation, and every JPEG-compressed or resized pair passes. The sample is selected by Python \texttt{Random(0).sample} from the sorted 47,238 reference JPG paths in class/filename order. Each transform acts independently on the original RGB image. For side fraction $f$, the crop has width $\lfloor fw\rfloor$ and height $\lfloor fh\rfloor$, with each offset given by half the remaining dimension rounded down; it is resized back to $(w,h)$. Rotations call \texttt{rotate(deg, fillcolor=(0,0,0))} at $+2^\circ,+5^\circ,+10^\circ$, without explicit resampling or expansion arguments. JPEG quality is set to 90, 70 or 50, then the result is decoded as RGB. Resize controls pass through $128\times128$ or $384\times384$ before returning to the original dimensions. Other resampling and JPEG encoder options use Pillow defaults; this diagnostic's Pillow and codec versions were not recorded. These pairs characterize the tested transformations and frames, not real-world duplicate recall.

\begin{table}[htbp]
\centering
\setlength{\tabcolsep}{2pt}
\arrayrulecolor{black!65}
\caption{Synthetic joint-rule hash sensitivity on 50 Kvasir-Capsule frames from 30 video prefixes. Each cell counts original--transformed pairs satisfying both 64-bit hash thresholds. Crop percentages indicate the retained fraction of each side; cropped and resized images return to the original dimensions. The same frames are used in every row. These pass rates describe the tested transformations and frames, not recall on naturally occurring duplicates.}
\label{tab:detection_limits}
\begin{tabular}{lrr}
\toprule[0.4pt]
\rowcolor{TableHeader} Transformation & $d_p,d_d\leq6$ & $d_p,d_d\leq2$ \\
\midrule[0.4pt]
Center crop, 90\% & \cellcolor{ClaimTint} 0/50 (0\%) & 0/50 (0\%) \\
Center crop, 75\% & 0/50 (0\%) & 0/50 (0\%) \\
Center crop, 60\% & 0/50 (0\%) & 0/50 (0\%) \\
Rotation, $2^\circ$ & 40/50 (80\%) & 5/50 (10\%) \\
Rotation, $5^\circ$ & 5/50 (10\%) & 0/50 (0\%) \\
Rotation, $10^\circ$ & 0/50 (0\%) & 0/50 (0\%) \\
JPEG, quality 90 & \cellcolor{ClaimTint} 50/50 (100\%) & 50/50 (100\%) \\
JPEG, quality 70 & 50/50 (100\%) & 50/50 (100\%) \\
JPEG, quality 50 & 50/50 (100\%) & 48/50 (96\%) \\
Resize, $128\times128$ & 50/50 (100\%) & 50/50 (100\%) \\
Resize, $384\times384$ & 50/50 (100\%) & 50/50 (100\%) \\
\bottomrule[0.4pt]
\end{tabular}
\end{table}

\FloatBarrier

\subsection{Within-CV2024 comparisons and same-modality controls}
Within the public split, Table~\ref{tab:internal_merged} counts validation rows with a same-source training counterpart under each stated predicate. Among the 1,381 exact-pHash KVASIR rows, 540 have NCC $\geq0.99$ with the first lexicographic counterpart and 799 satisfy that threshold with any exact-pHash counterpart. These counts answer different matching questions; neither is a detector-recall estimate. On the same 1,381 selected pairs, grayscale $256\times256$ bilinear processing gives 387 pairs with NCC $\geq0.99$, compared with 540 at $96\times96$ bicubic. For centered float32 arrays with $S_a=\sum_i a_i^2$ and $S_b=\sum_i b_i^2$, the alternate calculation uses $\sqrt{S_aS_b}$ as its denominator without additive epsilon, returning zero when that denominator is at most $10^{-9}$. The $96\times96$ calculation instead uses $\sqrt{S_a}\sqrt{S_b}+10^{-12}$. Resolution, interpolation and normalization arithmetic therefore change together. All 1,381 selected pairs share the same video prefix between their two images. Separately, minimizing the frame-index gap over same-video exact-pHash training candidates gives median one and interquartile range 1--6 over these 1,381 validation rows. The nearest same-video training frame over all 11,581 KVASIR validation rows has median one and interquartile range 1--1.

\begin{table}[htbp]
\centering
\setlength{\tabcolsep}{2pt}
\arrayrulecolor{black!65}
\caption{Within-source public train--validation candidates. Counts are validation rows with a training counterpart under each screen, including separate rows sharing a basename. The joint rule requires both hash distances to be at most six for the same pair. Candidate counts do not imply pixel verification; byte identities are reported separately.}
\label{tab:internal_merged}
\begin{tabular}{lrrrr}
\toprule[0.4pt]
\rowcolor{TableHeader} Source & \shortstack{Validation\\rows} & \shortstack{Basename\\matched rows} & \shortstack{Exact-pHash\\rows} & \shortstack{Joint-rule\\candidate rows} \\
\midrule[0.4pt]
KVASIR & 11,581 & 0 & 1,381 & 4,896 \\
SEE-AI & 4,289 & 174 & 218 & 264 \\
KID & 165 & 0 & 1 & 4 \\
AIIMS & 97 & 0 & 6 & 6 \\
\bottomrule[0.4pt]
\end{tabular}
\end{table}

Table~\ref{tab:internal_pixels} summarizes pixel checks of the other-source candidate pairs. KID has pixel-screen candidates despite having no byte-identical crossing. Candidate pairs, affected validation rows and identity groups remain different units; NCC on resized grayscale images is not file-byte equality.

\begin{table}[htbp]
\centering
\setlength{\tabcolsep}{2pt}
\arrayrulecolor{black!65}
\caption{NCC for selected within-source train--validation candidate pairs. The Rows column gives the validation size; the Pairs column counts one selected training counterpart per eligible row, using the lexicographically first full path satisfying $d_p,d_d\leq6$ on that pair. NCC uses grayscale $96\times96$ bicubic resizing; threshold columns count selected pairs. These conditional counts do not estimate true-duplicate prevalence or detector precision.}
\label{tab:internal_pixels}
\begin{tabular}{lrrrrrr}
\toprule[0.4pt]
\rowcolor{TableHeader} Source & Rows & Pairs & Mean NCC & $\geq.99$ & $\geq.95$ & $\geq.90$ \\
\midrule[0.4pt]
SEE-AI & 4,289 & 264 & 0.9831 & 220 & 228 & 244 \\
KID & 165 & 4 & 0.9391 & 2 & 2 & 3 \\
AIIMS & 97 & 6 & 1.0000 & 6 & 6 & 6 \\
\bottomrule[0.4pt]
\end{tabular}
\end{table}

\FloatBarrier

SSIM~\citep{wang2004image} is computed with scikit-image~\citep{vanderwalt2014skimage}, using the same grayscale geometry, data range 255, uniform $7\times7$ windows and sample covariance. Each comparison group contains 1,381 pairs. Controls exclude flagged pairs, exact basenames and exact-pHash matches. The adjacent control pairs an unflagged validation frame with the eligible same-video training frame having the smallest absolute numeric frame-index gap; equal-gap ties follow the stored cache order. Candidate pairs are deduplicated, shuffled and subsampled to 1,381 using a shared random-number stream with seed 42. Distant same-video controls instead require a frame-index gap of at least 500. Table~\ref{tab:ssim} compares selected pairs with these controls.

The selected pairs have higher median SSIM than the controls, but the distributions overlap and 68 adjacent controls also reach NCC $\geq0.99$. Unflagged pairs are therefore comparisons, not verified negative examples.

Both images share a prefix in every selected, adjacent and distant pair. These inventories contain 37, 40 and 39 prefixes, respectively. For the adjacent and distant comparisons, we jointly resample their unions of 40 and 39 prefixes, applying the same multiplicities to selected and control pairs. The 95\% percentile intervals for selected-minus-control median SSIM are $[0.186,0.285]$ and $[0.401,0.487]$, respectively, over 10,000 draws, conditional on the pair inventories. These uncertainty intervals differ from the 5th--95th percentiles of individual SSIM values in Table~\ref{tab:ssim}. Different-video controls depend on the videos of both images, so grouping only by the validation image does not represent this dependence; their distributions are reported without confidence intervals. The non-KVASIR control spans only three source clusters and is likewise reported without a joint interval. None of these comparisons establishes different-patient sampling.

\begin{table}[htbp]
\centering
\setlength{\tabcolsep}{2pt}
\arrayrulecolor{black!65}
\caption{Structural similarity for selected pairs and comparison groups. Each group contains 1,381 pairs. The 5th--95th percentiles describe individual pair SSIM values. The 95\% joint-prefix intervals concern selected-minus-control median SSIM for same-video controls; they resample shared video prefixes together, conditional on the selected pairs. Dashes mark the reference row or comparisons without a reported joint interval.}
\label{tab:ssim}
\begin{tabular}{lrrrr}
\toprule[0.4pt]
\rowcolor{TableHeader} Pair group & $n$ & Median & \shortstack{5th--95th\\percentile} & \shortstack{95\% joint-prefix\\gap interval} \\
\midrule[0.4pt]
Selected exact-pHash & 1,381 & \cellcolor{ClaimTint} 0.880 & \cellcolor{ClaimTint} $[0.738,0.963]$ & --- \\
Adjacent same-video & 1,381 & \cellcolor{ClaimTint} 0.647 & \cellcolor{ClaimTint} $[0.313,0.899]$ & \cellcolor{ClaimTint} $[+0.186,+0.285]$ \\
Distant same-video & 1,381 & 0.432 & $[0.244,0.653]$ & $[+0.401,+0.487]$ \\
Different-video, same-class & 1,381 & 0.442 & $[0.246,0.615]$ & --- \\
Different-video, any-class & 1,381 & 0.441 & $[0.253,0.612]$ & --- \\
Non-KVASIR, same-source/class & 1,381 & 0.330 & $[0.191,0.495]$ & --- \\
\bottomrule[0.4pt]
\end{tabular}
\end{table}

\subsection{Sensitivity to hypothetical video grouping}
\label{app:clusters}
The exact-pHash rate $1,381/11,581=11.9247\%$ is a census of the KVASIR validation list. Table~\ref{tab:clusters} describes sensitivity to hypothetical prefix-resampling assumptions, without estimating patient membership or patient-level uncertainty. Each draw computes total flagged rows divided by total rows, preserving resampling multiplicity rather than averaging prefix rates equally.

Each scenario draws $m$ prefixes with replacement from the 41-prefix empirical distribution, using 20,000 draws and resetting NumPy's default generator to seed 42. We report raw percentile intervals for 41, 30, 20 and ten sampled prefixes, with widths relative to the 41-prefix interval. The ten-prefix scenario is the widest tested, not a patient-level bound; unknown patient groups could be fewer in number, unequal in size or otherwise different from these sampled prefixes.

\begin{table}[htbp]
\centering
\setlength{\tabcolsep}{2pt}
\arrayrulecolor{black!65}
\caption{Sensitivity to the number of resampled video prefixes. The exact-pHash rate, 11.9247\%, is a census of KVASIR validation. These conditional percentile ranges describe hypothetical prefix sampling, not uncertainty in the enumerated count or a patient-level bound.}
\label{tab:clusters}
\begin{tabular}{lrr}
\toprule[0.4pt]
\rowcolor{TableHeader} Sampled prefixes & 95\% range (\%) & Width / 41-prefix width \\
\midrule[0.4pt]
41 & $[6.899,17.996]$ & 1.000 \\
30 & $[6.353,19.046]$ & 1.144 \\
20 & $[5.636,20.807]$ & 1.367 \\
10 & $[4.389,24.824]$ & 1.841 \\
\bottomrule[0.4pt]
\end{tabular}
\end{table}

A complementary scheme partitions all 41 prefixes into synthetic groups with balanced prefix counts, then resamples those groups. Across 1,000 random partitions and 20 resamples per partition, the ranges for 30, 20 and ten groups are $\BalancedCoarseningThirtyRange\%$, $\BalancedCoarseningTwentyRange\%$ and $\BalancedCoarseningTenRange\%$, respectively, compared with $\BalancedCoarseningFortyOneRange\%$ for 41 prefixes. The largest shift of any interval limit is $\BalancedCoarseningMaxShiftPP$ percentage points. Balanced coarsening retains all prefixes in each partition; drawing fewer prefixes addresses a different hypothetical sampling target. Neither construction identifies the unknown patient grouping or guarantees its uncertainty coverage.

\section{Fixed-prediction diagnostics and official-test evaluation}
\label{app:rescore}
\subsection{Public-submission diagnostics}
\label{app:public_predictions}
We join submitted predictions by full path and named class columns, then re-score them on fixed evaluation views. The challenge score averages class-macro AUC and BA~\citep{capsule_vision_2024}; AUC uses the supplied scores without an added softmax. The 27 workbooks use aliases A--AA. Scoreability requires exact full-path alignment to all 16,132 validation rows, all ten named class columns, no missing values and no duplicate paths. Entries A--Y pass; Z contains bare filenames, 44 of which occur more than once (45 surplus rows), while AA contains absolute Windows paths that do not match the required paths. Both fail the path check; no path repair is applied. Of the 25 scoreable entries, 24 reproduce the organizer-reported full-list AUC, BA and combined score with absolute errors below $10^{-12}$, before display rounding. Team G's workbook re-scores at BA 0.100 and AUC 0.378, whereas its reported full-list values are 0.899 and 0.944. Table~\ref{tab:m7} subtracts the reported full-list score, so the all-25 contrast includes this discrepancy as well as row filtering. The primary, score-consistent 24-entry cohort excludes Team G. The 19-entry subset additionally excludes Teams U, V, W, X and Y. In an earlier version of this work, this subset served as the headline cohort on the premise that its teams had trained their own models; that premise was not independently verified, so we retain the subset only to show sensitivity to cohort membership. An 18-entry sensitivity also omits Team A, which scores 1.0 on every public view. Such a score is consistent with predictions from a model trained on public validation, which the challenge permitted for final-test models (next paragraph); it does not by itself indicate a rule violation. The aliases keep team names out of the text but do not anonymize entries: the release pins each public workbook, whose filename names its team, by SHA-256.

Final-test training could use both public partitions. The supplied records do not establish, for every array, whether public-validation rows were excluded from fitting or whether the same checkpoint generated the official-test predictions. Re-scoring therefore describes fixed prediction records, without certifying held-out performance.

\begin{table}[htbp]
\centering
\setlength{\tabcolsep}{2pt}
\arrayrulecolor{black!65}

\caption{Public-submission re-scoring. Score differences are relative to the organizer-reported public-validation score, the mean of macro AUC and balanced accuracy. Nominal 95\% bias-corrected and accelerated (BCa) ranges resample entries individually with frames fixed. Entry independence is unestablished; we report descriptive resampling ranges and do not interpret them as calibrated population intervals. Rank changes are within each cohort. The all-25 rows use a mixed baseline: one reported public-validation score disagrees with full-list re-scoring. The primary cohort excludes that entry; the fixed subset additionally omits Teams U--Y, and its sensitivity further omits Team A.}
\label{tab:m7}

\begin{tabular}{llrrrr}
\toprule[0.4pt]
\rowcolor{TableHeader} Evaluation view & Cohort & $n$ & Mean $\Delta$ & \shortstack{Nominal 95\%\\BCa} & \shortstack{Entries\\changing rank} \\
\midrule[0.4pt]
Non-KVASIR & All (mixed baseline) & 25 & $-$0.073 & $[-0.167, -0.043]$ & 18 \\
Non-KVASIR & Primary & 24 & $-$0.050 & $[-0.067, -0.032]$ & 19 \\
Non-KVASIR & Subset & 19 & $-$0.062 & $[-0.079, -0.045]$ & 14 \\
Non-KVASIR & Subset excl. A & 18 & $-$0.066 & $[-0.082, -0.049]$ & 14 \\
Internal-filtered & All (mixed baseline) & 25 & $-$0.063 & $[-0.165, -0.034]$ & 18 \\
Internal-filtered & Primary & 24 & $-$0.039 & $[-0.056, -0.023]$ & 16 \\
Internal-filtered & Subset & 19 & $-$0.049 & $[-0.066, -0.034]$ & 11 \\
Internal-filtered & Subset excl. A & 18 & $-$0.051 & $[-0.069, -0.037]$ & 11 \\
\bottomrule[0.4pt]
\end{tabular}

\end{table}

The nominal 95\% BCa ranges~\citep{efron1987bca} resample entries independently with evaluation frames fixed. SciPy uses 10,000 resamples per table row and one advancing NumPy default-generator stream initialized at seed 42, in table-row order. Within each view, entries enter the resampling in descending order of unrounded combined score, and each cohort keeps that order. Because independence among entries has not been established, we do not interpret these ranges as calibrated confidence intervals for a population of participants or models. Filtering does not improve public-to-official rank agreement in any of the four cohorts. These are evaluation-view diagnostics of stored predictions, not estimates of the benefit of retraining.

Equal-source averaging changes the weighting and represented-class support of the metric. Within a source, BA averages represented-class recall and AUC uses classes for which the one-versus-rest comparison is defined; the combined score averages those two quantities. We then average the source scores equally. For the primary 24-entry cohort, the non-KVASIR change in equal-source combined score is $\ScoreConsistentEqualNonKV$, compared with $\ScoreConsistentPooledNonKV$ under pooled aggregation (Table~\ref{tab:rescore_sensitivity}, which repeats the pooled and equal-source combined changes of Table~\ref{tab:main_evidence}(d) and adds equal-source BA). The 19-entry sensitivity gives $-0.037$ in equal-source BA and $-0.022$ in equal-source combined score. These comparisons make the metric and cohort dependence explicit. Small KID/AIIMS slices and unequal class support limit comparisons beyond these fixed lists.

\begin{table}[htbp]
\centering
\setlength{\tabcolsep}{2pt}
\arrayrulecolor{black!65}
\caption{Mean paired score changes (view minus public validation) for the primary 24-entry score-consistent cohort (Team G alone excluded) and the fixed 19-entry subset sensitivity (Teams U--Y additionally omitted). The pooled challenge score averages BA and macro AUC. Equal-source scores weight sources equally: four in the public view and three in the non-KVASIR and filtered views. Within each source, BA averages recalls over present classes, and the combined score averages BA and present-class macro AUC. These descriptive means hold predictions fixed.}
\label{tab:rescore_sensitivity}
\begin{tabular}{llrr}
\toprule[0.4pt]
\rowcolor{TableHeader} Metric & Cohort ($n$) & Non-KVASIR & Internal-filtered \\
\midrule[0.4pt]
Pooled challenge score & 24 (primary) & \cellcolor{ClaimTint} $-$0.0503 & \cellcolor{ClaimTint} $-$0.0393 \\
Equal-source BA & 24 (primary) & $-$0.0308 & $-$0.0255 \\
Equal-source combined & 24 (primary) & \cellcolor{ClaimTint} $-$0.0199 & \cellcolor{ClaimTint} $-$0.0163 \\
\midrule[0.4pt]
Pooled challenge score & 19 (subset) & $-$0.0624 & $-$0.0488 \\
Equal-source BA & 19 (subset) & $-$0.0370 & $-$0.0305 \\
Equal-source combined & 19 (subset) & $-$0.0220 & $-$0.0177 \\
\bottomrule[0.4pt]
\end{tabular}
\end{table}

Table~\ref{tab:rank_alignment} reports the descriptive association with official-test ranks using each cohort's own entries. Team A, whose public score is 1.0, strongly affects the 19-entry coefficients: excluding it raises them from 0.244, 0.068 and 0.139 to 0.463, 0.257 and 0.340 on the public, non-KVASIR and internal-filtered views. Spearman correlations use unrounded combined scores and average ranks for exact ties; no exact score ties occur in these cohorts. Lower observed correlation on the reduced views does not establish a population-level difference between correlations. These results do not calibrate either diagnostic view as a replacement leaderboard.

\begin{table}[htbp]
\centering
\setlength{\tabcolsep}{2pt}
\arrayrulecolor{black!65}
\caption{Spearman correlation of public-view combined scores with organizer-reported official-test combined scores. The Public column uses the organizer-reported scores; the non-KVASIR and internal-filtered columns use recalculated scores. The primary 24-entry score-consistent cohort excludes only Team G, whose reported public-validation score disagrees with re-scoring. The fixed 19-entry subset sensitivity additionally omits Teams U--Y, without an independently justified exclusion rationale, and the 18-entry sensitivity also omits Team A. All four cohorts have lower observed agreement on both reduced views; possible dependence among entries limits population-level inference.}
\label{tab:rank_alignment}
\begin{tabular}{lrrrr}
\toprule[0.4pt]
\rowcolor{TableHeader} Cohort & $n$ & Public & Non-KVASIR & Internal-filtered \\
\midrule[0.4pt]
Score-consistent (primary) & 24 & \cellcolor{ClaimTint} 0.557 & \cellcolor{ClaimTint} 0.459 & \cellcolor{ClaimTint} 0.497 \\
All scoreable & 25 & 0.566 & 0.355 & 0.390 \\
19-entry subset sensitivity & 19 & 0.244 & 0.068 & 0.139 \\
18-entry subset excl.\ A & 18 & 0.463 & 0.257 & 0.340 \\
\bottomrule[0.4pt]
\end{tabular}
\end{table}

\FloatBarrier

\subsection{Official AIIMS test evaluation}
\label{app:official_aiims}
A separate byte scan of the 4,385 released official-test files finds no match with the public training folder, one label-disjoint match with public validation and eight within-test duplicate groups. This scan describes released file content; it establishes neither participant-specific exposure nor patient independence.

The separate official-test study evaluates DINOv2-L/14 head/LoRA adaptation on full, random-10,596 and source-only-10,596 training pools at epoch ten over ten seeds. Unlike the two-class public AIIMS subset, the official AIIMS test represents all ten classes. Its 4,385-row population and source-only training pool are distinct from the public-validation 10,591-row removal comparison. Table~\ref{tab:official} reports means and standard errors, extending Table~\ref{tab:main_evidence}(c) with the combined score and accuracy. Normal makes up 80.4\% of the test rows, and all three pools have lower accuracy than always predicting Normal; BA weights the ten classes equally. The release includes the producer code, each run's arguments, GPU model and library versions, and the recorded confusion matrices and class-specific AUCs, permitting re-aggregation of BA, accuracy and mean AUC (Appendix~\ref{app:reproducibility}). The producer saved no checkpoints or per-image predictions, so checks beyond these aggregates require retraining.

\begin{table}[htbp]
\centering
\small
\setlength{\tabcolsep}{2pt}
\arrayrulecolor{black!65}

\caption{Official AIIMS test evaluation. Mean $\pm$ standard error (SE) over ten seeds. Source-only training uses 10,596 rows, distinct from the 10,591-row removal input. The combined score averages balanced accuracy and mean AUC. Paired by seed, Source-only exceeds Full in mean AUC in all ten seeds, whereas the interval for its BA difference includes zero (see text); these results do not validate either filtered public view as an official-test proxy.}
\label{tab:official}

\begin{tabular}{lrrrr}
\toprule[0.4pt]
\rowcolor{TableHeader} Training pool & \shortstack{Balanced\\accuracy} & Mean AUC & Combined & Accuracy \\
\midrule[0.4pt]
Full & \cellcolor{ClaimTint} $0.2804\pm0.0046$ & \cellcolor{ClaimTint} $0.7177\pm0.0015$ & $0.4991\pm0.0026$ & $0.1758\pm0.0098$ \\
Random-10,596 & $0.2786\pm0.0038$ & $0.7352\pm0.0036$ & $0.5069\pm0.0028$ & $0.3142\pm0.0119$ \\
Source-only-10,596 & \cellcolor{ClaimTint} $0.2688\pm0.0034$ & \cellcolor{ClaimTint} $0.7533\pm0.0013$ & $0.5110\pm0.0019$ & $0.0988\pm0.0063$ \\
\bottomrule[0.4pt]
\end{tabular}

\end{table}

Paired by seed label, Source-only minus Full is $+0.0356$ in mean AUC (95\% paired-$t$ interval $[+0.0309,+0.0403]$; positive in all ten seeds) and $-0.0117$ in BA ($[-0.0259,+0.0026]$; negative in seven). The AUC interval excludes zero, whereas the BA interval does not; the combined score, which averages the two, changes by $+0.0120$ ($[+0.0032,+0.0207]$). Against Random, Source-only is $+0.0180$ in AUC ($[+0.0112,+0.0249]$) and $-0.0098$ in BA ($[-0.0230,+0.0033]$). Seed labels pair the runs, not their GPU models. \texttt{replay.py} recomputes these contrasts from the released confusion matrices and class AUCs. These results do not calibrate the non-KVASIR or internal-filtered public views as replacements for official-test evaluation.

\section{Training design and evaluation populations}
\label{app:controls}
Training-pool summaries use $\pm$ for sample standard deviation unless labeled SE or $h_{95}$. In Table~\ref{tab:main_evidence}, $h_{95}$ denotes the half-width of the paired-$t$ interval, not a standard error. For paired differences $d_s$, the 95\% interval is $\bar d\pm t_{9,0.975}s_d/\sqrt{10}$, assuming independent, approximately normal seed-wise contrasts. Seed-percentile intervals use 10,000 paired resamples, NumPy's default generator with seed 42 and linearly interpolated quantiles. These pointwise intervals are conditional on fixed lists and pool construction. They provide neither simultaneous coverage over all displayed contrasts nor patient-sampling uncertainty. Means and interval limits are rounded separately.

\subsection{Optimization and training pools}
\label{app:optimization}
DINOv2-L/14 uses a LayerNorm--linear--GELU--dropout--linear classification head with 256 hidden units and dropout probability 0.1. Rank-eight LoRA modules with scale $16/8$ adapt the combined query/key/value projections in the final four blocks. For each adapted linear map, the added term is $2W_{\mathrm{up}}W_{\mathrm{down}}x$, with entries of $W_{\mathrm{down}}$ initialized independently as $\mathcal{N}(0,0.01^2)$ and $W_{\mathrm{up}}$ initialized to zero. Backbone evaluation mode is maintained during training. The head trains throughout ten epochs; LoRA updates begin after three head-only epochs. AdamW~\citep{loshchilov2019adamw} retains its state across epochs and uses a head learning rate of $3\times10^{-4}$, a LoRA learning rate of $10^{-4}$ and weight decay of $10^{-4}$. We use the default betas $(0.9,0.999)$ and epsilon $10^{-8}$, with no learning-rate schedule. Cross-entropy uses label smoothing 0.05 and effective-number class weights~\citep{cui2019classbalancedloss} $w_c=10r_c/\sum_jr_j$, where $r_c=(1-0.999)/(1-0.999^{n_c})$ for a class with $n_c$ training rows; every primary pool contains all ten classes.

Training and evaluation use deterministic direct $224\times224$ resizing and ImageNet normalization. The training implementation applies TorchVision \texttt{Resize((224,224))} to PIL RGB inputs: bilinear interpolation with PIL antialiasing under TorchVision 0.21.0. Both stages use FP16 autocast; evaluation softmax uses float32 logits. Models are initialized through the Torch Hub entry \texttt{dinov2\_vitl14} in the local DINOv2 cache, without an explicit weight override. The repository revision and per-run frozen-backbone digests were not recorded. Training batches contain 128 rows, with shuffled order and the final incomplete batch dropped. Full training supplies 293 minibatches per epoch, compared with 82 for each removal, source-only and size-matched control pool. The eleven DINOv2 conditions share this pipeline, A100 hardware, seeds 0--9 and evaluation at epoch ten, giving 110 model fits. Within each condition, the training list and hyperparameters are fixed across seeds. Pool-construction experiments considered seeds 0 and 42; Appendix~\ref{app:composition} specifies the retained sampling rules. The available records do not fully resolve the original hyperparameter and list-selection rationale; inference remains conditional on the retained settings and lists.

The \removalpool{} pool excludes all KVASIR training rows and five other-source rows positive under the hash screen, leaving 10,591 rows. The five additional exclusions comprise two SEE-AI Erosion images, one SEE-AI Polyp image, one KID Normal image and one AIIMS Worms image. \sourcepool{} excludes only KVASIR and contains 10,596 rows. Table~\ref{tab:dedup_le6_breakdown} gives their class/source support; Table~\ref{tab:training_pools} reports every training-pool condition on public validation.

\begin{table}[htbp]
\centering
\setlength{\tabcolsep}{2pt}
\arrayrulecolor{black!65}

\caption{Class and source composition after source removal. Source-only training contains 10,596 rows; non-KVASIR validation contains 4,551 rows. The last column gives class totals for the distinct 10,591-row Removal input.}
\label{tab:dedup_le6_breakdown}

\begin{tabular}{lrrrrrrr}
\toprule[0.4pt]
\rowcolor{TableHeader}  & \multicolumn{3}{c}{\shortstack{Source-only\\training}} & \multicolumn{3}{c}{\shortstack{Non-KVASIR\\validation}} & Removal \\
\cmidrule(lr){2-4}\cmidrule(lr){5-7}
\rowcolor{TableHeader} Class & SEE-AI & KID & AIIMS & SEE-AI & KID & AIIMS & training \\
\midrule[0.4pt]
Angioectasia & 530 & 18 & 0 & 228 & 9 & 0 & 548 \\
Bleeding & 519 & 3 & 0 & 223 & 2 & 0 & 522 \\
Erosion & 2,340 & 0 & 0 & 1,003 & 0 & 0 & 2,338 \\
Erythema & 580 & 0 & 0 & 249 & 0 & 0 & 580 \\
Foreign Body & 249 & 0 & 0 & 107 & 0 & 0 & 249 \\
Lymphangiectasia & 376 & 6 & 0 & 162 & 3 & 0 & 382 \\
Normal & 4,312 & 315 & 0 & 1,849 & 136 & 0 & 4,626 \\
Polyp & 1,090 & 34 & 0 & 468 & 15 & 0 & 1,123 \\
Ulcer & 0 & 0 & 66 & 0 & 0 & 29 & 66 \\
Worms & 0 & 0 & 158 & 0 & 0 & 68 & 157 \\
Total & 9,996 & 376 & 224 & 4,289 & 165 & 97 & 10,591 \\
\bottomrule[0.4pt]
\end{tabular}

\end{table}

\subsection{Matched composition controls}
\label{app:composition}
The random control draws 10,596 rows from full training. Class-match-R is a separate 10,591-row draw matching the class counts of \removalpool{}; Comp contains 10,596 rows and matches the class counts of \sourcepool{}. Ordered lists are reproducible from the training workbook without presorting or sampling with replacement. Random uses pandas~\citep{pandas2010} \texttt{sample(n=10596, random\_state=0)} and retains the returned order. For both class-matched pools, process classes in Table~\ref{tab:dedup_le6_breakdown} order: draw as many KVASIR rows as possible up to each target count, then fill any remainder from non-KVASIR. Comp uses the integer \texttt{random\_state=0} separately for every draw. Class-match-R instead passes one advancing NumPy \texttt{RandomState(0)} through all draws. Concatenate class blocks with KVASIR rows first, then shuffle once using a fresh integer \texttt{random\_state=0}. These recipes reproduce every ordered row with pandas 2.1.4 and NumPy 1.26.4. Both class-matched pools preferentially sample KVASIR; they are not uniform class-only draws. Comp contains 6,687 KVASIR, 3,716 SEE-AI, 158 AIIMS and 35 KID rows. Its Normal, Angioectasia, Foreign Body, Lymphangiectasia and Ulcer rows are all KVASIR; its Erosion, Erythema and Polyp rows are 81--94\% SEE-AI; its Worms rows are all AIIMS; and its Bleeding rows are 312 KVASIR, 209 SEE-AI and one KID. Class-match-R differs from these counts by at most two rows in any class-by-source cell. Pool sampling seeds are distinct from training seeds.

The following controls are variants of Comp and use the same training pipeline. Comp contains 66 KVASIR Ulcer images. Whenever Normal rows are replaced below, the same 66 KVASIR Normal rows are used, selected from Comp in its existing order with pandas \texttt{sample(n=66, random\_state=42)}.
\begin{description}
\item[A.] Replace Comp's 66 KVASIR Ulcer images with 66 AIIMS Ulcer images, preserving every non-Ulcer identity.
\item[B.] Starting from Comp, add those 66 AIIMS Ulcer images and remove the 66 Normal rows; retain the original KVASIR Ulcer images.
\item[C.] Starting from A, replace the 66 Normal rows with second copies of its 66 AIIMS Ulcer images.
\item[$D_{\mathrm{dup}}$.] Starting from Comp, replace the 66 Normal rows with second copies of its 66 KVASIR Ulcer images.
\item[$D_{\mathrm{uniq}}$.] Starting from Comp, replace the 66 Normal rows with 66 additional unique KVASIR Ulcer images.
\end{description}
For $D_{\mathrm{uniq}}$, exclude Comp's Ulcer paths from Full's KVASIR Ulcer rows, retain workbook order, then apply \texttt{sample(n=66, random\_state=42)}. In A--D, retained rows keep their relative order and additions are appended without a final shuffle. Added AIIMS rows follow workbook order; duplicated KVASIR Ulcer rows follow Comp order. Thus A preserves non-Ulcer identities and relative order, not their absolute positions. B, C and both D pools share the same remaining background. Table~\ref{tab:training_pools} reports all training pools before the paired comparisons in Table~\ref{tab:control_contrasts}.

\begin{table}[htbp]
\centering
\setlength{\tabcolsep}{2pt}
\arrayrulecolor{black!65}
\caption{Training-pool results on the public validation list (16,132 rows). Balanced accuracy is mean $\pm$ sample SD over ten seeds at epoch ten. Removal and Class-match-R contain 10,591 rows; other reduced pools contain 10,596. Replacement controls A--D are defined in Appendix~\ref{app:composition}.}
\label{tab:training_pools}
\begin{tabular}{lrrr}
\toprule[0.4pt]
\rowcolor{TableHeader} Training pool & Rows & Balanced accuracy & $\Delta$ vs. Full \\
\midrule[0.4pt]
Full & 37,607 & $0.825\pm0.008$ & --- \\
Random & 10,596 & $0.779\pm0.005$ & $-$0.046 \\
Source-only & 10,596 & $0.611\pm0.008$ & $-$0.214 \\
Removal & 10,591 & $0.612\pm0.007$ & $-$0.213 \\
\addlinespace
Class-match-R & 10,591 & $0.738\pm0.005$ & $-$0.087 \\
Comp & 10,596 & $0.734\pm0.007$ & $-$0.092 \\
A & 10,596 & $0.665\pm0.003$ & $-$0.161 \\
B & 10,596 & $0.743\pm0.006$ & $-$0.082 \\
C & 10,596 & $0.661\pm0.005$ & $-$0.165 \\
D$_{\mathrm{dup}}$ & 10,596 & $0.737\pm0.005$ & $-$0.088 \\
D$_{\mathrm{uniq}}$ & 10,596 & $0.739\pm0.005$ & $-$0.086 \\
\bottomrule[0.4pt]
\end{tabular}
\end{table}

Table~\ref{tab:control_contrasts} compares these specified list changes on their common backgrounds. A-minus-Comp measures replacement of the specified Ulcer images; C-minus-A additionally changes class counts and repetition. These comparisons do not exhaust class/source interactions, and their sums are arithmetic decompositions rather than identified causal shares of the full removal effect.

\begin{table}[htbp]
\centering
\setlength{\tabcolsep}{2pt}
\arrayrulecolor{black!65}
\caption{Training-pool contrasts by evaluation population. Each cell gives the mean BA difference and a pointwise 95\% paired-$t$ interval over ten seeds. Each contrast includes any changes to class weights and sampling induced by the stated training-pool changes.}
\label{tab:control_contrasts}
\begin{tabular}{lrr}
\toprule[0.4pt]
\rowcolor{TableHeader} Contrast & Public $\Delta$ [95\% CI] & Non-KVASIR $\Delta$ [95\% CI] \\
\midrule[0.4pt]
Source-only $-$ Random & \cellcolor{ClaimTint} \shortstack{$-0.1682$\\$[-0.1754,-0.1610]$} & \cellcolor{ClaimTint} \shortstack{$+0.0624$\\$[+0.0516,+0.0732]$} \\
\addlinespace
A $-$ Comp & \cellcolor{ClaimTint} \shortstack{$-0.0690$\\$[-0.0752,-0.0628]$} & \cellcolor{ClaimTint} \shortstack{$+0.0978$\\$[+0.0898,+0.1057]$} \\
\addlinespace
B $-$ Comp & \shortstack{$+0.0092$\\$[+0.0034,+0.0151]$} & \shortstack{$+0.0927$\\$[+0.0855,+0.0999]$} \\
\addlinespace
C $-$ A & \shortstack{$-0.0042$\\$[-0.0085,+0.0002]$} & \shortstack{$-0.0050$\\$[-0.0111,+0.0011]$} \\
\addlinespace
D$_{\mathrm{dup}}$ $-$ C & \shortstack{$+0.0765$\\$[+0.0706,+0.0824]$} & \shortstack{$-0.0882$\\$[-0.0961,-0.0802]$} \\
\addlinespace
D$_{\mathrm{dup}}$ $-$ Comp & \shortstack{$+0.0033$\\$[-0.0022,+0.0088]$} & \shortstack{$+0.0046$\\$[-0.0026,+0.0118]$} \\
\addlinespace
D$_{\mathrm{uniq}}$ $-$ D$_{\mathrm{dup}}$ & \shortstack{$+0.0020$\\$[+0.0010,+0.0030]$} & \shortstack{$+0.0002$\\$[-0.0010,+0.0014]$} \\
\addlinespace
Removal $-$ Source-only & \cellcolor{ClaimTint} \shortstack{$+0.0011$\\$[-0.0044,+0.0066]$} & \cellcolor{ClaimTint} \shortstack{$+0.0017$\\$[-0.0078,+0.0112]$} \\
\bottomrule[0.4pt]
\end{tabular}
\end{table}

\FloatBarrier

For C-minus-A on public validation, the 95\% seed-percentile interval $\CMinusAPercentileCI$ excludes zero, whereas the paired-$t$ interval in Table~\ref{tab:control_contrasts} includes it. Evidence for a nonzero contrast therefore depends on the interval construction; neither result establishes equivalence.

The separate equivalence check for B-minus-Comp uses a 90\% paired-$t$ interval around the $+0.009238$ BA difference: $\FreshBCompCI$. The $\pm0.010$ margin was fixed in an earlier version of this work, after the first four seeds of a previous run of this comparison; the fits reported here were trained later, and the margin has no external calibration. Because the 90\% interval extends beyond $+0.010$, equivalence within the margin is not established. Likewise, Removal-minus-Source-only is $\FiveRowDelta$ with 95\% interval $\FiveRowCI$. This contrast measures the response to deleting five rows under the training pipeline, including the resulting changes to weights and sampling; an interval spanning zero does not demonstrate equivalence or isolate an image-content mechanism.

\subsection{Validation views, class support and uncertainty}
Table~\ref{tab:training_views} crosses training pools with three fixed evaluation views: public, non-KVASIR and internal-filtered validation. It repeats the values of Table~\ref{tab:main_evidence}(a) so that the per-source and per-class results below can be read beside them. The predictions within each training condition are held fixed across views, separating evaluation-population sensitivity from retraining. All ten classes occur in every view.

\begin{table}[htbp]
\centering
\setlength{\tabcolsep}{2pt}
\arrayrulecolor{black!65}
\caption{DINOv2-L/14 training pools crossed with fixed validation views. Entries are epoch-ten mean BA over ten matched seeds; predictions are held fixed across views. Public, non-KVASIR and internal-filtered views contain 16,132, 4,551 and 4,326 rows, respectively, each supporting all ten classes.}
\label{tab:training_views}
\begin{tabular}{lrrrr}
\toprule[0.4pt]
\rowcolor{TableHeader} Training pool & Train rows & Public & Non-KVASIR & Internal-filtered \\
\midrule[0.4pt]
Full & 37,607 & \cellcolor{ClaimTint} 0.825 & \cellcolor{ClaimTint} 0.745 & \cellcolor{ClaimTint} 0.773 \\
Random & 10,596 & 0.779 & 0.695 & 0.707 \\
Comp & 10,596 & 0.734 & 0.388 & 0.401 \\
Class-match-R & 10,591 & 0.738 & 0.391 & 0.406 \\
Removal & 10,591 & \cellcolor{ClaimTint} 0.612 & \cellcolor{ClaimTint} 0.759 & \cellcolor{ClaimTint} 0.778 \\
Source-only & 10,596 & 0.611 & 0.757 & 0.776 \\
\bottomrule[0.4pt]
\end{tabular}
\end{table}

The internal filter starts with the 4,551 non-KVASIR validation rows. All 37,607 training and 16,132 validation rows have pHash values, so the implementation's hash-availability condition excludes none. Within each source, validation rows are flagged if their basename occurs in training or their pHash equals a training pHash. The union of flagged basenames across sources removes every matching validation row, preserving order: 225 rows with 221 unique basenames are removed, leaving 4,326.

\begin{table}[htbp]
\centering
\setlength{\tabcolsep}{2pt}
\arrayrulecolor{black!65}
\caption{Validation performance by source. BA averages recall over the classes represented in each source. The contrast is Removal minus Full; intervals use ten paired seeds. Source-wise macro averages are not additive contributions to pooled BA.}
\label{tab:source_metrics}
\begin{tabular}{lrrrrrr}
\toprule[0.4pt]
\rowcolor{TableHeader} Source & $n$ & Classes & Full & Removal & $\Delta$ & 95\% paired-$t$ \\
\midrule[0.4pt]
KVASIR & 11,581 & 9 & 0.943 & 0.470 & $-$0.4730 & $[-0.4817,-0.4644]$ \\
SEE-AI & 4,289 & 8 & 0.697 & 0.710 & +0.0129 & $[+0.0013,+0.0245]$ \\
KID & 165 & 5 & 0.812 & 0.768 & $-$0.0443 & $[-0.0863,-0.0022]$ \\
AIIMS & 97 & 2 & 0.929 & 0.951 & +0.0215 & $[+0.0083,+0.0346]$ \\
\bottomrule[0.4pt]
\end{tabular}
\end{table}

Removal-minus-Full BA reverses sign, becoming positive on the 4,551-row non-KVASIR view. Source-only-minus-Random is also positive there despite being negative on public validation. Table~\ref{tab:source_metrics} reports source-specific BA over the classes represented in each source. KVASIR lacks Worms; SEE-AI lacks Ulcer and Worms. KID represents Angioectasia, Bleeding, Lymphangiectasia, Normal and Polyp, while public AIIMS represents Ulcer and Worms.

The 4,326-row internal-filtered view further excludes within-source candidate matches. Its Removal-minus-Full contrast is $\InternalMainDelta$ ($\InternalMainCI$), and its Source-only-minus-Full contrast is $\InternalSourceDelta$ ($\InternalSourceCI$). Both intervals span zero. Filtering defines diagnostic populations and does not establish patient independence.

\begin{table}[htbp]
\centering
\setlength{\tabcolsep}{2pt}
\arrayrulecolor{black!65}
\caption{Per-class mean recall on public validation over ten seeds. All classes are reported. Removal-minus-Full recall differences average to the pooled BA contrast.}
\label{tab:class_recall}
\begin{tabular}{lrrrr}
\toprule[0.4pt]
\rowcolor{TableHeader} Class & Rows & Full & Removal & $\Delta$ recall \\
\midrule[0.4pt]
Angioectasia & 497 & 0.789 & 0.667 & $-$0.122 \\
Bleeding & 359 & 0.855 & 0.842 & $-$0.014 \\
Erosion & 1,155 & 0.704 & 0.657 & $-$0.047 \\
Erythema & 297 & 0.558 & 0.403 & $-$0.155 \\
Foreign Body & 340 & 0.896 & 0.650 & $-$0.246 \\
Lymphangiectasia & 343 & 0.825 & 0.592 & $-$0.233 \\
Normal & 12,287 & 0.985 & 0.587 & $-$0.397 \\
Polyp & 500 & 0.664 & 0.636 & $-$0.029 \\
Ulcer & 286 & 0.976 & 0.096 & $-$0.879 \\
Worms & 68 & 1.000 & 0.991 & $-$0.009 \\
\bottomrule[0.4pt]
\end{tabular}
\end{table}

\FloatBarrier

The class-count-matched pools drawn from Full have lower non-KVASIR BA than Full: Comp and Class-match-R reach 0.388 and 0.391 BA, respectively, compared with 0.745 for Full. Thus a control can match class counts without preserving performance on another source mixture. Because the dominant source differs across classes in these pools (Appendix~\ref{app:composition}), source appearance can predict class; the non-KVASIR loss is consistent with this shortcut. The class-wise recall differences in Table~\ref{tab:class_recall} average to the pooled BA difference. The large Ulcer change accompanies strong source/class association and sparse non-KVASIR support, so it cannot be attributed to class-specific content alone.

On public validation, Full and Removal have mean AUCs $\TrainingFullAUC$ and $\TrainingRemovalAUC$, and combined scores $\TrainingFullCombined$ and $\TrainingRemovalCombined$ (change $\TrainingCombinedDelta$). Selecting each run's highest public-validation BA gives $\TrainingFullBestBA$ and $\TrainingRemovalBestBA$, a contrast of $\TrainingBestDelta$ compared with $\TrainingFinalDelta$ at epoch ten.

\subsection{Frozen-backbone CNN comparisons}
ResNet-50 and ConvNeXt-Tiny use frozen ImageNet-pretrained backbones held in evaluation mode and the same 256-unit nonlinear head structure described above. TorchVision weight identifiers are \texttt{IMAGENET1K\_V2} and \texttt{IMAGENET1K\_V1}, respectively. Their input transform resizes the short side to 256 with bilinear interpolation and antialiasing enabled, center-crops to 224 and applies ImageNet normalization. Training uses FP16 autocast for ResNet-50 and BF16 for ConvNeXt-Tiny; evaluation runs without autocast. AdamW uses learning rate $3\times10^{-4}$, weight decay $10^{-4}$, the same class-weighted loss and batch size 128. Cosine decay spans the ten-epoch update budget. Each backbone is trained with ten seeds on full training, source-only training and the matched random pool and evaluated at epoch ten, giving 60 model fits. Within each seed, conditions use the same physical A100. BatchNorm buffers remain fixed, stochastic depth is inactive, and the schedule advances after successful mixed-precision updates.

\begin{table}[htbp]
\centering
\setlength{\tabcolsep}{2pt}
\arrayrulecolor{black!65}
\caption{Source sensitivity with frozen CNN backbones and nonlinear heads. Full uses 37,607 training rows; Random and Source-only each use 10,596. Values are ten-seed mean BA; $\Delta$ is Source-only minus Random.}
\label{tab:cnn_final}
\begin{tabular}{lrrrrr}
\toprule[0.4pt]
\rowcolor{TableHeader} Backbone & Validation & Full & Random & \shortstack{Source-\\only} & $\Delta$ \\
\midrule[0.4pt]
ResNet-50 & Public & 0.697 & 0.658 & 0.458 & \cellcolor{ClaimTint} $-$0.2002 \\
ResNet-50 & Non-KVASIR & 0.578 & 0.522 & 0.610 & \cellcolor{ClaimTint} +0.0878 \\
ConvNeXt-Tiny & Public & 0.692 & 0.645 & 0.436 & \cellcolor{ClaimTint} $-$0.2096 \\
ConvNeXt-Tiny & Non-KVASIR & 0.562 & 0.507 & 0.602 & \cellcolor{ClaimTint} +0.0955 \\
\bottomrule[0.4pt]
\end{tabular}
\end{table}

Both CNNs have negative Source-only-minus-Random BA contrasts on public validation and positive contrasts on non-KVASIR validation (Table~\ref{tab:cnn_final}). For ResNet-50, the corresponding 95\% paired-$t$ intervals are $\ResNetPublicCI$ and $\ResNetNonKVCI$; for ConvNeXt-Tiny, they are $\ConvNeXtPublicCI$ and $\ConvNeXtNonKVCI$. Thus, the direction of this contrast depends on the evaluation population in both frozen-backbone configurations as well as the adapted transformer. Their absolute scores do not provide a controlled architecture comparison because adaptation and preprocessing differ.

\section{Sensitivity to removal rules and supervision}
\label{app:sensitivity_controls}
These controls test how removing source-attributed CV2024 training rows or changing their labels affects performance. They use the DINOv2-L/14 head/LoRA architecture, adapter initialization, image transform, loss and optimizer settings specified in Appendix~\ref{app:optimization}. Class weights follow Appendix~\ref{app:optimization} for training lists with all ten classes; AIIMS removal lacks Worms training support, as discussed below.

The threshold, label and source-omission families use ten seed identities and final-epoch metrics on fixed evaluation lists; the matched add-back comparison uses four. GPU models differ across the source-omission and threshold experiments, so common seed labels do not imply matched hardware. At each epoch, these runs also evaluate Kvasir-Capsule's \texttt{split\_1} (Appendix~\ref{app:auxiliary}), mapped to CV2024 labels, without using it to select the reported epoch.

These results were aggregated from per-seed final-epoch records; complete per-image predictions are unavailable for these auxiliary runs. They are descriptive controls for the stated pipelines, not isolated causal contributions to the removal effect.

\subsection{Hash and pixel removal thresholds}
\label{app:threshold_controls}
The pHash/dHash controls apply the hash screen, with its independent minima, at radii zero and two. The NCC controls remove selected source matches above the stated pixel cutoff. Both families remove only KVASIR training rows and retain all 10,596 other-source rows. Table~\ref{tab:threshold_controls} reports their training populations. Lowering a pixel cutoff removes more rows, changing size, class counts and source content together. The NCC $\geq0.80$ arm retains two KVASIR rows as well as the five other-source rows absent from Removal, so the difference from Removal is seven training rows.

The hash arms use L40S GPUs, the NCC arms use A100 or RTX A6000 GPUs, and Source-only uses RTX A6000; Full and Removal use A100. Absolute scores are therefore not a common-hardware comparison with the primary results. The family's Full, Removal and Source-only references are its own run records, distinct from the primary fits in Appendix~\ref{app:controls}; their means nevertheless agree with the primary means to three decimals.

\begin{table}[htbp]
\centering
\setlength{\tabcolsep}{2pt}
\arrayrulecolor{black!65}
\caption{Training-data removal thresholds for DINOv2-L/14, evaluated on all 16,132 public-validation rows. BA is the epoch-ten mean $\pm$ sample SD over seeds 0--9. $d_p$ and $d_d$ are independently minimized pHash and dHash distances to the source reference; NCC uses the selected source counterpart. Full, Source-only and Removal are this family's own reference runs. Appendix~\ref{app:threshold_controls} describes these auxiliary experiments and their hardware differences.}
\label{tab:threshold_controls}
\begin{tabular}{lrrr}
\toprule[0.4pt]
\rowcolor{TableHeader} Removal rule & Train rows & KVASIR kept & Public BA \\
\midrule[0.4pt]
Full & 37,607 & 27,011 & $0.825\pm0.008$ \\
$d_p,d_d\le 0$ & 21,241 & 10,645 & $0.813\pm0.007$ \\
$d_p,d_d\le 2$ & 10,825 & 229 & $0.712\pm0.008$ \\
NCC $\ge 0.99$ & 12,697 & 2,101 & $0.744\pm0.008$ \\
NCC $\ge 0.95$ & 10,715 & 119 & $0.699\pm0.004$ \\
NCC $\ge 0.90$ & 10,613 & 17 & $0.654\pm0.008$ \\
NCC $\ge 0.85$ & 10,605 & 9 & $0.618\pm0.008$ \\
NCC $\ge 0.80$ & 10,598 & 2 & $0.615\pm0.010$ \\
Source-only & 10,596 & 0 & $0.611\pm0.008$ \\
Removal & 10,591 & 0 & $0.612\pm0.007$ \\
\bottomrule[0.4pt]
\end{tabular}
\end{table}

Within the threshold-control training family, the BA loss of the NCC $\geq0.90$ arm relative to the family's Full reference is approximately 80\% of the Removal loss. This percentage is an arithmetic comparison of fixed-list outcomes, not a decomposition of leakage. Stricter cutoffs need not yield monotonic changes in source-wise performance.

\FloatBarrier
\subsection{Label assignment and source removal}
The reassignment condition uses ten fixed seed-specific lists. Each preserves all 37,607 image paths and all 10,596 non-KVASIR labels, but changes all 27,011 KVASIR labels to different CV2024 classes. All ten classes occur as replacement labels, and the KVASIR histogram changes. The sampling procedure for these lists is unavailable, so results are conditional on those assignments. The permutation control instead permutes the KVASIR one-hot label rows, preserving the image list and class histogram. This permutation still lowers public BA to approximately 0.504 (Table~\ref{tab:label_controls}). Initialization and label corruption share a seed, so seed variation includes both choices. This contrast establishes sensitivity to supervision while leaving instance memorization unidentified.

\begin{table}[htbp]
\centering
\setlength{\tabcolsep}{2pt}
\arrayrulecolor{black!65}
\caption{Sensitivity to KVASIR label assignment. Each model retains all 37,607 training paths; only the labels of the 27,011 KVASIR rows change. BA is mean $\pm$ sample SD at epoch ten over ten seeds.}
\label{tab:label_controls}
\begin{tabular}{lrrr}
\toprule[0.4pt]
\rowcolor{TableHeader} Label control & Histogram kept & Public BA & Non-KVASIR BA \\
\midrule[0.4pt]
Reassignment & No & $0.414\pm0.006$ & $0.716\pm0.010$ \\
Permutation & Yes & $0.504\pm0.010$ & $0.736\pm0.011$ \\
\bottomrule[0.4pt]
\end{tabular}
\end{table}

Removing SEE-AI or KID also lowers performance on the corresponding held-out source (Table~\ref{tab:source_removal_controls}). These experiments differ from scoring the KVASIR-removal models separately by source: each changes the training source itself. Class support differs across these experiments, and there is no matched-size control for each removed source. Source-omission runs use A100, RTX A6000 or L40S GPUs. AIIMS removal eliminates the sole Worms training support and is therefore not treated as a directly comparable source-generalization test. The results support a broader source-dependence interpretation without separating domain familiarity from image-level exposure.

\begin{table}[htbp]
\centering
\setlength{\tabcolsep}{2pt}
\arrayrulecolor{black!65}
\caption{Held-out-source BA after removing that source from training. Each row is a distinct training intervention; evaluation uses only the omitted source's public-validation rows. Full is the corresponding full-training reference from each experiment's own runs, so its KVASIR value (0.942) differs in the third decimal from the primary runs in Table~\ref{tab:source_metrics} (0.943). Omitted BA reports epoch-ten mean $\pm$ sample SD over ten seeds.}
\label{tab:source_removal_controls}
\begin{tabular}{lrrrrr}
\toprule[0.4pt]
\rowcolor{TableHeader} Omitted source & Train rows & Eval. rows & Classes & Full BA & Omitted BA \\
\midrule[0.4pt]
KVASIR & 10,596 & 11,581 & 9 & \cellcolor{ClaimTint} 0.942 & \cellcolor{ClaimTint} $0.466\pm0.015$ \\
SEE-AI & 27,611 & 4,289 & 8 & \cellcolor{ClaimTint} 0.697 & \cellcolor{ClaimTint} $0.239\pm0.008$ \\
KID & 37,231 & 165 & 5 & 0.812 & $0.495\pm0.057$ \\
\bottomrule[0.4pt]
\end{tabular}
\end{table}

\subsection{Related-source add-back}
This control asks how much public BA returns when a small set of Kvasir-Capsule frames is restored to Source-only. Adding 284 Kvasir-Capsule rows to the 10,596-row Source-only pool raises public BA from $\TrainingAddbackComparatorBA$ to $\TrainingAddbackBA$ across four matched seeds (change $\TrainingAddbackDelta$). The baseline and add-back arms share the DINOv2-L/14 head and rank-eight LoRA configuration, training script, optimizer settings, seeds 0--3, batch size 128, ten epochs and RTX A6000 hardware. The add-back arm also ran seeds 4--9, four on RTX A6000 and two on A100, without that paired baseline. Over all ten seeds it reaches mean public BA 0.703. Paired by seed label with the A100 Removal reference runs of Appendix~\ref{app:threshold_controls}, it is higher by $+0.091$ (95\% paired-$t$ interval $[+0.084,+0.099]$; positive in all ten seeds). That pool difference is 289 rows, the 284 added frames plus the five hash-screen rows that Removal excludes, and eight of the ten pairs mix GPU models. All added filenames already occur in CV2024 (200 in training and 84 in validation), and all are included in the related-source \texttt{split\_1} evaluation list. The construction excludes designated nearest-reference filenames, not every possible occurrence or match.

The result therefore describes re-exposure under a changed training pool. Because sample size, class weights and content all change, it does not measure unseen-source transfer or an isolated source-domain effect.

\section{Local versus distant removal: probability and accuracy}
\label{app:intervention}
\subsection{Matched lists, training and estimand}
Let $E_0$ be the initial 799 \emph{exposed} validation rows, each with at least one exact-pHash training counterpart reaching NCC $\geq0.99$. Before testing stratum eligibility, define the common removal set as
\[ R_0=\{t:\exists v\in E_0,\ d_p(v,t)=0\ \land\ \operatorname{NCC}(v,t)\geq0.99\}. \]
The 1,715 qualifying pairs have 1,100 distinct training rows. Each removed row must belong to a pair that individually passes both conditions; expanding to all exact-pHash counterparts of $E_0$ would instead remove 1,563 rows. Every arm removes $R_0$, unchanged by subsequent eligibility filtering.

Eligibility requires enough distant training rows to match a stratum's proposed local removal. One stratum requires 256 distant controls but has only 141 eligible rows, so its 211 exposed evaluation rows are excluded. Its 298 qualifying training counterparts remain in the common removal set, while its proposed local neighbors remain in every arm's training list. Additional local and distant removals come only from eligible strata.

Within each video/class training list sorted by frame index and filename, local removal takes the immediately preceding and following positions around each removed counterpart, merges overlapping neighborhoods, and excludes the counterpart positions. Distant controls match the lengths of the resulting local runs with nonoverlapping windows drawn uniformly from currently valid windows, using seeded restarts. Eligible distant rows lie outside the local set and at least 50 frame indices from every exposed row in the stratum. They belong to no cross-split exact-pHash group and have no exposed counterpart for which both hash distances are at most six. Local rows have a median frame-index gap of two to the nearest exposed row and a 95th percentile of eight; six have a gap of at least 50. List adjacency therefore need not imply temporal proximity. The eligible strata supply 876 additional removals per arm, giving 1,976 removals in total.

Before eligibility filtering, unflagged validation rows were sampled without replacement to match the 799 exposed rows by video/class. After excluding the ineligible stratum from both groups, evaluation retains 588 exposed and 588 unflagged rows. Unflagged rows have no pHash in a cross-split exact-pHash group. Each evaluation group contains 547 Normal, 19 Angioectasia, 15 Bleeding, four Erythema, two Foreign Body and one Lymphangiectasia row. Every minority class occurs in only one prefix. Conclusions concern these eligible groups with uneven class support.

The four arms are local removal and three fixed distant-control draws, numbered 101, 202 and 303. They use the DINOv2-L/14 head/LoRA architecture, preprocessing, class-weighted objective, optimization settings and batch size in Appendix~\ref{app:controls}, with ten training epochs. The ten seeds are 4--9 and 14--17. All four arms within a seed use the same GPU model; seed blocks run on NVIDIA A100, RTX A6000 or L40S devices. The 40 model fits yield 47,040 epoch-ten prediction rows, including repeated evaluation of the same frames. The release includes these probabilities, the fixed evaluation and removal lists and the training code (Appendix~\ref{app:reproducibility}). Equation~\ref{eq:did} first contrasts local removal with the mean of the three distant arms within each seed and evaluation group, then subtracts the unflagged contrast from the exposed contrast. It compares removal strategies and does not identify an exposure-versus-no-exposure or memorization effect.

\subsection{Conditional inference and design scope}
The design, seed set, control choices and endpoint criterion were selected with access to evaluation outcomes. Seeds 14--17 were run first, after earlier designs with different arms (one on these same seeds); seeds 4--9 were added later. Apart from one-epoch smoke runs, the retained records contain no other seed for the reported design. All intervals are exploratory, conditional on the selected lists and unadjusted for selection. Their 95\% levels are nominal; coverage after outcome-informed design and endpoint selection is not established. The three distant-control draws remain fixed across seeds.

The endpoints are true-class probability, negative log-likelihood with probability floor $10^{-12}$, true-class probability minus the largest other-class probability, micro accuracy and macro recall over the six represented classes. Probability, margin and accuracy changes are expressed in percentage points where indicated. The probability contrast is $-\ProbDropPP$ percentage points, with paired-$t$ interval $[\ProbCILoPP,\ProbCIHiPP]$. The NLL difference-in-differences is positive and the margin difference-in-differences is negative across all ten seeds. These directions describe correlated changes in the same probability vectors, not independent confirmations. Accuracy and BA each have four negative and six zero seed contrasts, with paired-$t$ intervals spanning zero; the experiment does not establish an accuracy decrease.

Table~\ref{tab:local_endpoints} in the main text gives absolute endpoint levels and separates the two components of Eq.~\ref{eq:did}. The exposed local-minus-distant probability change is $\LocalExposedProbabilityPP$ percentage points, compared with $\LocalUnflaggedProbabilityPP$ for matched-unflagged rows. The exposed component is negative in eight seeds and positive in two; its difference from the unflagged component is negative in all ten. Thus ten negative differences-in-differences do not imply a negative exposed-group contrast in every seed. All four absolute levels compare removal strategies; none is an unremoved-model baseline. Video/class matching does not equalize the observed endpoint levels.

Table~\ref{tab:confidence} gives paired-$t$ intervals, restating the $h_{95}$ half-widths of Table~\ref{tab:main_evidence}(e) as intervals and adding NLL and the probability margin; Table~\ref{tab:confidence_uncertainty} gives bootstrap ranges. Seed resampling preserves all four arms within a seed. Prefix resampling jointly weights the 32 contributing prefixes across arms and evaluation groups while averaging the ten fixed seeds. Both resampling procedures use 10,000 draws. They condition on different components and are not independent confirmations; neither jointly integrates seed randomness, control selection and unknown patient membership. For accuracy and BA, seed-percentile intervals exclude zero, whereas the reported prefix ranges reach zero and the paired-$t$ intervals span it.

The fixed-six-class BA statistic is undefined in 9,086 of 10,000 prefix-bootstrap draws (90.86\%); only 914 retain all six classes. We therefore do not report a percentile interval for that fixed-class target under this resampling rule. The reported present-class BA range instead averages recall over the classes present in each draw, a different, draw-dependent target. Omitting absent classes does not recover fixed-six-class inference. Continuous endpoints and micro accuracy do not require this absent-class averaging rule. The 32 leave-one-prefix-out mean probability contrasts range from $\LocalProbabilityLOPOMinPP$ to $\LocalProbabilityLOPOMaxPP$ percentage points. The three largest-magnitude signed prefix contributions sum to $\LocalTopThreePrefixSharePercent\%$ of the signed net contrast. Normal contributes approximately 68\%; these overlapping decompositions cannot be added.

\begin{table}[htbp]
\centering
\setlength{\tabcolsep}{2pt}
\arrayrulecolor{black!65}
\caption{Local-versus-distant removal: exposed-minus-unflagged contrasts in Eq.~\ref{eq:did}. Probability, probability-margin, balanced-accuracy and accuracy contrasts are in absolute score units, not percentage points; NLL uses natural logarithms. Seed signs count negative/zero/positive contrasts over ten paired seeds. Intervals are pointwise and do not adjust for design selection.}
\label{tab:confidence}
\begin{tabular}{lrrr}
\toprule[0.4pt]
\rowcolor{TableHeader} Endpoint & Mean contrast & 95\% paired-$t$ interval & Seed signs \\
\midrule[0.4pt]
True-class probability & \cellcolor{ClaimTint} $-$0.00355 & \cellcolor{ClaimTint} $[-0.00486,-0.00224]$ & 10/0/0 \\
NLL & +0.00495 & $[+0.00329,+0.00660]$ & 0/0/10 \\
Probability margin & $-$0.00472 & $[-0.00647,-0.00297]$ & 10/0/0 \\
\addlinespace
Balanced accuracy & \cellcolor{ClaimTint} $-$0.00194 & \cellcolor{ClaimTint} $[-0.00485,+0.00097]$ & 4/6/0 \\
Accuracy & \cellcolor{ClaimTint} $-$0.00079 & \cellcolor{ClaimTint} $[-0.00163,+0.00004]$ & 4/6/0 \\
\bottomrule[0.4pt]
\end{tabular}
\end{table}

\begin{table}[htbp]
\centering
\setlength{\tabcolsep}{2pt}
\arrayrulecolor{black!65}
\caption{Bootstrap ranges for local-versus-distant removal: exposed-minus-unflagged, local-minus-distant contrasts in Eq.~\ref{eq:did}. Probability, margin, balanced-accuracy and accuracy contrasts are in absolute score units, not percentage points; NLL uses natural logarithms. Ranges are pointwise and exploratory, without adjustment for design selection. Seed resampling retains the pairing of all four arms. Video-prefix resampling is joint across arms and evaluation strata, with training seeds averaged within each draw. For balanced accuracy, the seed range uses the fixed six classes, whereas the prefix range averages recall over the classes present in each draw.}
\label{tab:confidence_uncertainty}
\begin{tabular}{lrr}
\toprule[0.4pt]
\rowcolor{TableHeader} Endpoint & 95\% seed percentile & \shortstack{95\% video-prefix\\percentile range} \\
\midrule[0.4pt]
True-class probability & $[-0.00462,-0.00244]$ & $[-0.00693,-0.00067]$ \\
NLL & $[+0.00357,+0.00632]$ & $[+0.00092,+0.00931]$ \\
Probability margin & $[-0.00615,-0.00325]$ & $[-0.00920,-0.00093]$ \\
Balanced accuracy & $[-0.00460,-0.00004]$ & $[-0.00399,+0.00000]$ \\
Accuracy & $[-0.00153,-0.00017]$ & $[-0.00161,+0.00000]$ \\
\bottomrule[0.4pt]
\end{tabular}
\end{table}

\FloatBarrier

\section{Kvasir-Capsule lists and training-population sensitivities}
\label{app:kvasir_protocols}
The upstream lists define a separate evaluation setting from CV2024. We first state their overlap properties, then describe the grouped-split, frame-count and video-selection experiments that use them.

\subsection{Official Kvasir-Capsule list scope}
\label{app:auxiliary}
The official Kvasir-Capsule lists~\citep{smedsrud2021} \texttt{split\_0} and \texttt{split\_1} contain 23,061 and 24,100 rows, respectively, at the inspected repository revision.\footnote{\raggedright \texttt{official\_splits/split\_0.csv} and \texttt{split\_1.csv} at revision \texttt{553601dc316f}: \url{https://github.com/simula/kvasir-capsule/tree/553601dc316fef82f39c952e28f60aaad66039e2/official_splits}, independently rechecked on 28 October 2026.} Each has 25 video prefixes, with 43 in their union and seven shared prefixes. These lists differ from the 47,238-entry class-folder inventory used for source matching; list-row totals are not counts of unique reference images. The second list has 24,092 unique filenames; the lists have no filename intersection or shared video/class pair, and the minimum cross-fold frame-index gap among frames from the same video is 201. Filename disjointness thus coexists with prefix overlap, without establishing adjacent same-class reuse or patient disjointness. The seven shared prefixes in these distributed lists differ from the video-disjoint description in \citet{smedsrud2021} (Technical Validation); this observation does not establish which lists produced that article's experimental results. These boundaries belong to Kvasir-Capsule, not the CV2024 public split or the AIIMS test; their overlap properties cannot be transferred between protocols.

\subsection{Kvasir-Capsule split and training-population sensitivities}
\label{app:kvasir_auxiliary}
These experiments concern the upstream Kvasir-Capsule lists, rather than the ten-class CV2024 task. They illustrate why evaluation boundaries, represented-class support and epoch selection must accompany a reported score, using video grouping, training-set size and video selection as distinct design choices. The tables were generated from \KvasirAuxRuns{} run histories, with recorded counts and class supports checked against the split metadata.

\paragraph{Training procedure and recorded epochs.}
The models use DINOv2-L/14 with rank-eight LoRA on the last four attention blocks and a 256-unit nonlinear head. Inputs are resized directly to $224\times224$ with ImageNet normalization. Training uses batches of 128, class-weighted cross-entropy with label smoothing 0.05 and head/LoRA learning rates $3\times10^{-4}/10^{-4}$. Class weights follow the effective-number rule of Appendix~\ref{app:optimization} over the eleven classes, with $n_c$ counted in the training list; a class absent from that list enters the normalizing sum with $r_c=0.001$ and then receives weight zero. Unlike the primary pipeline, these runs use two rather than three head-only epochs, recreate AdamW each epoch and place the backbone in training mode during optimization. Epoch caps are 20 for grouped splitting and scaling and 12 for video selection, with evaluation-BA patience of five and four epochs, respectively. Terminal results use the last recorded epoch; selected results use the epoch with the highest evaluation BA in each history. Because evaluation controls stopping and selection, neither value is an untouched-test estimate. Means and sample standard deviations describe the two recorded runs per setting.

\paragraph{Video-grouped folds.}
The five folds partition the video prefixes in the combined official lists. BA averages recall over the \KvasirGroupTestClassMin{}--\KvasirGroupTestClassMax{} classes present in each test fold. The Blood class is absent from training in one fold. The two recorded runs per fold reset the training random generators to zero, so their labels do not denote independent initialization seeds. Their descriptive spread therefore cannot quantify seed uncertainty; overlapping training folds also preclude treating fold scores as independent dataset replications. Table~\ref{tab:kvasir_grouped} reports outcomes for each fold's represented classes.

\begin{table}[htbp]
\centering
\setlength{\tabcolsep}{2pt}
\arrayrulecolor{black!65}
\caption{Video-grouped evaluation on Kvasir-Capsule (zero-based folds). BA is mean $\pm$ descriptive across-record SD over two recorded runs. Terminal denotes the last recorded epoch; best is selected using evaluation BA. Evaluation BA also determines early stopping. Last recorded epochs: 10--17. BA averages recall over the classes present in each test fold; training sets overlap across folds.}
\label{tab:kvasir_grouped}
\begin{tabular}{lrrrrr}
\toprule[0.4pt]
\rowcolor{TableHeader} Fold & \shortstack{Test\\rows} & \shortstack{Test\\classes} & \shortstack{Missing train\\class} & \shortstack{Terminal\\BA} & \shortstack{Selected best\\BA} \\
\midrule[0.4pt]
0 & 9,418 & 8 & None & $0.3959 \pm 0.0002$ & $0.4127 \pm 0.0000$ \\
1 & 9,421 & 9 & None & $0.2668 \pm 0.0003$ & $0.2869 \pm 0.0003$ \\
2 & 9,425 & 6 & None & $0.4105 \pm 0.0013$ & $0.4805 \pm 0.0001$ \\
3 & 9,475 & 9 & None & $0.3922 \pm 0.0025$ & $0.4392 \pm 0.0001$ \\
4 & 9,422 & 8 & Blood & $0.2726 \pm 0.0000$ & $0.2801 \pm 0.0005$ \\
\bottomrule[0.4pt]
\end{tabular}
\end{table}

\paragraph{Training-set size.}
Class-stratified samples of the official \texttt{split\_0} list are evaluated on \texttt{split\_1}, which contains all eleven classes. Sampling seeds change the selected rows, while training resets its random generators to zero. The full-size condition consequently repeats the same list and initialization. Table~\ref{tab:kvasir_scaling} separates the two recorded values. Mean terminal BA is \KvasirScaleHalfTerminal{} at half size and \KvasirScaleFullTerminal{} at full size in these recorded runs. Row count, update count and stopping epoch vary together; these runs do not identify a general scaling law.

\begin{table}[htbp]
\centering
\setlength{\tabcolsep}{2pt}
\arrayrulecolor{black!65}
\caption{Class-stratified training-frame sampling on Kvasir-Capsule. BA is mean $\pm$ descriptive across-record SD over two recorded runs. Terminal denotes the last recorded epoch; best is selected using evaluation BA. Evaluation BA also determines early stopping.}
\label{tab:kvasir_scaling}
\begin{tabular}{lrrrr}
\toprule[0.4pt]
\rowcolor{TableHeader} Fraction & \shortstack{Train\\rows} & \shortstack{Terminal\\BA} & \shortstack{Selected best\\BA} & \shortstack{Last recorded\\epoch range} \\
\midrule[0.4pt]
0.1 & 2,302 & $0.1339 \pm 0.0175$ & $0.1666 \pm 0.0057$ & 7--11 \\
0.25 & 5,761 & $0.1846 \pm 0.0206$ & $0.2123 \pm 0.0057$ & 8--11 \\
0.5 & 11,528 & $0.2364 \pm 0.0084$ & $0.2501 \pm 0.0125$ & 7--9 \\
1.0 & 23,061 & $0.2339 \pm 0.0001$ & $0.2574 \pm 0.0000$ & 11 \\
\bottomrule[0.4pt]
\end{tabular}
\end{table}

\paragraph{Video selection at a fixed row budget.}
The final experiment draws different numbers of \texttt{split\_0} videos and constructs 23,000 training rows by cyclic repetition or random truncation of their frames. Evaluation remains on \texttt{split\_1}. Here the seed controls both video selection and training initialization. The selected frame pool and class coverage change with the video count; some training pools lack classes present in evaluation. The official lists also share video prefixes (Appendix~\ref{app:auxiliary}). Mean terminal BA increases from \KvasirDiversityLowTerminal{} for the smallest video set to \KvasirDiversityHighTerminal{} for the largest (Table~\ref{tab:kvasir_diversity}). This is an association under the stated sampling and stopping procedure, not an isolated video-count effect or evidence of patient-disjoint generalization.

\begin{table}[htbp]
\centering
\setlength{\tabcolsep}{2pt}
\arrayrulecolor{black!65}
\caption{Kvasir-Capsule video selection at a fixed budget of 23,000 effective training rows. BA is mean $\pm$ descriptive across-record SD over two recorded runs. Terminal denotes the last recorded epoch; best is selected using evaluation BA. Evaluation BA also determines early stopping. Pool size precedes repetition or truncation to the effective training list.}
\label{tab:kvasir_diversity}
\begin{tabular}{lrrrr}
\toprule[0.4pt]
\rowcolor{TableHeader} Videos & \shortstack{Selected pool\\rows (mean)} & \shortstack{Terminal\\BA} & \shortstack{Selected best\\BA} & \shortstack{Last recorded\\epoch range} \\
\midrule[0.4pt]
5 & 4,255.5 & $0.1835 \pm 0.0417$ & $0.1987 \pm 0.0398$ & 5--7 \\
10 & 8,744.0 & $0.1939 \pm 0.0251$ & $0.2313 \pm 0.0009$ & 9 \\
15 & 14,009.0 & $0.2094 \pm 0.0162$ & $0.2244 \pm 0.0120$ & 9--12 \\
20 & 17,289.0 & $0.2117 \pm 0.0076$ & $0.2401 \pm 0.0017$ & 5--8 \\
25 & 23,061.0 & $0.2400 \pm 0.0079$ & $0.2640 \pm 0.0009$ & 10 \\
\bottomrule[0.4pt]
\end{tabular}
\end{table}

\FloatBarrier

\section{External screens and their scope}
\label{app:external}
For ChestX-ray14, we inspect \texttt{train\_val\_list.txt} and \texttt{test\_list.txt}, acquired from the \texttt{alkzar90/NIH-Chest-X-ray-dataset} mirror of the NIH release. Rechecking a fixed mirror revision reproduces the counts below; the original acquisition date and revision were not retained.\footnote{\raggedright NIH provider distribution: \url{https://nihcc.app.box.com/v/ChestXray-NIHCC}. Lists rechecked on 27 October 2026 at mirror revision \texttt{36778e3b0e4f}: \url{https://huggingface.co/datasets/alkzar90/NIH-Chest-X-ray-dataset/tree/36778e3b0e4f4b4fad31d1728d6190f3eda5b543/data}.} The lists contain 86,524 and 25,596 filenames, with 28,008 and 2,797 patient prefixes. The filename and patient-prefix intersections are both empty. This complete-list check concerns identifiers in the release associated with the ChestX-ray dataset family~\citep{wang2017chestxray}; pixel content and the correctness of patient identifiers are not independently checked.

The image-level screens use the 25,331-image ISIC 2019 training archive~\citep{isic2019data,tschandl2018ham10000,codella2018isic2017,hernandezperez2024bcn20000} and 10,662 labeled HyperKvasir images~\citep{borgli2020hyperkvasir}. ISIC comparisons cross metadata strata within that inventory. BCN, HAM and MSK denote prefixes of the metadata field \texttt{lesion\_id}; empty IDs form the unknown-lesion-ID stratum. These are metadata groups, not verified independent institutions. For HyperKvasir, we compare its labeled inventory with CV2024 KVASIR images. Neither comparison estimates train--validation prevalence, and the latter does not audit HyperKvasir's cross-validation folds.

\begin{table}[htbp]
\centering
\setlength{\tabcolsep}{2pt}
\arrayrulecolor{black!65}

\caption{External hash-candidate pairs and pixel similarity. The sum-rule and joint-rule columns score single pairs under different rules. Every joint-rule candidate was evaluated from original images. ISIC strata follow lesion-ID metadata, including an unknown-ID group; they are not verified institutions. NCC columns count joint-rule candidates passing the threshold using $96\times96$ bicubic and $256\times256$ bilinear grayscale images.}
\label{tab:external_pixels}

\begin{tabular}{lrrrr}
\toprule[0.4pt]
\rowcolor{TableHeader} Comparison & $d_p+d_d\leq6$ & $d_p,d_d\leq6$ & NCC $\geq.99$, 96 & NCC $\geq.99$, 256 \\
\midrule[0.4pt]
BCN / HAM & 0 & 1 & 0 & 0 \\
HAM / unknown lesion ID & 1 & 64 & 0 & 0 \\
Unknown lesion ID / MSK & 2 & 2 & 2 & 2 \\
Other ISIC combinations & 0 & 0 & 0 & 0 \\
ISIC total & 3 & 67 & 2 & 2 \\
HyperKvasir / CV2024 KVASIR & 0 & 1 & 0 & 0 \\
\bottomrule[0.4pt]
\end{tabular}

\end{table}

The sum rule ($d_p+d_d\leq6$) yields three ISIC candidates and no HyperKvasir/CV2024 KVASIR candidate. The broader joint rule ($d_p\leq6$ and $d_d\leq6$) yields 67 and one, respectively (Table~\ref{tab:external_pixels}). Both rules score a single pair, unlike the hash screen's independent minima. We evaluate every joint-rule candidate. Two ISIC pairs pass NCC $\geq0.99$ and the HyperKvasir pair fails under both $96\times96$ bicubic and $256\times256$ bilinear grayscale processing. Byte comparison establishes identity for one passing ISIC pair; similarity alone does not establish it for the other. Both link an empty lesion ID to an MSK-prefixed ID. These pair counts estimate neither image-level prevalence nor detector specificity.

These checks cover ReuScope's identifier and image-screening components. Pair counts here, affected validation-row counts and the cross-release removals reported by \citet{cassidy2022isic} use different populations and units. Detector-recall comparisons also require aligned identifiers, annotations and search rules; a shared NCC threshold is insufficient. An empty candidate set does not exclude other forms of overlap.

\section{Reproducibility, access and asset terms}
\label{app:reproducibility}
\subsection{Code and data access}
The ReuScope repository at \url{https://github.com/eunseobchoi/ReuScope} provides the audit code, the training and inference code of the primary, local-removal, CNN and official-test runs, the released data and scripts that recompute reported results from them. Its \texttt{REPRODUCE.md} maps each released result to its producing code, inputs, outputs and checking command, and lists the analyses whose code is not released. Its \texttt{data/results} folder (5\,MB) contains ordered primary training-pool and evaluation memberships; fixed local evaluation and removal lists; retained byte-identity, video-prefix and pair measurements; primary/CNN seed-level metric records; 47,040 local prediction rows with ten-class probabilities; official-test confusion matrices and class-specific AUCs; and derived public-entry scores. The \texttt{data/predictions} folder (127\,MB) contains the saved epoch-ten probabilities of all 110 primary and 60 CNN fits on their evaluation lists. A Croissant 1.1 description (\texttt{croissant.json}) lists every released data file with its SHA-256.

The list representation preserves row order and repeated occurrences, with image paths relative to the public dataset root; an exporter writes the training and evaluation CSVs read by the reported runs, with identical rows and order. Participant submission workbooks are read from the challenge repository at a fixed revision rather than redistributed. Original images and pretrained weights remain with their providers. Model checkpoints, per-epoch histories (which give the best-epoch values in Section~\ref{sec:counterfactual} and Appendix~\ref{app:controls}), execution logs, code for the auxiliary training analyses (Appendix~\ref{app:sensitivity_controls}) and Kvasir-Capsule experiments (Appendix~\ref{app:kvasir_protocols}), and the code of the feature, transform, PDQ, video-prefix resampling and external-release screens (Appendix~\ref{app:external}) are not distributed; Appendix~\ref{app:numerical} lists every omitted code component.

\subsection{Computational environment and compute}
Primary DINOv2 and CNN runs record one visible A100 (80\,GB) per fit. Local-removal runs configure one A100 (80\,GB), RTX A6000 (48\,GB) or L40S (48\,GB) per seed. Additional CV2024 controls ran on these GPU models, as identified in the run records, without recorded device counts. Official-test runs used L40S (28 fits), A100 and RTX A6000 GPUs; the launcher requested one GPU per run and two A100s for Full seed 0, for which the producer uses data parallelism when several devices are visible, and the visible device count was not recorded. Kvasir-Capsule runs recorded neither GPU model nor count. Jobs ran on a shared university GPU cluster. Primary training uses PyTorch 2.6.0~\citep{pytorch2024preferred}, CUDA 12.4 and cuDNN 9.1. Contemporaneous environment probes and launch records identify TorchVision~\citep{torchvision2024} \texttt{0.21.0+cu124} and Pillow~\citep{pillow} \texttt{12.1.1} for the primary campaign; these are campaign-level records, not per-fit version fields. CNN runs record Python 3.10.18, NumPy 2.2.6, pandas 2.3.3 and scikit-learn~\citep{pedregosa2011scikit} 1.7.2. The released code computes AUCs with scikit-learn's \texttt{roc\_auc\_score}. Numerical re-aggregation used Python 3.12.3, NumPy 1.26.4~\citep{harris2020array}, SciPy 1.17.1~\citep{scipy2020}, pandas 2.1.4 and scikit-learn 1.9.0, without loading model weights. Image-processing and CNN settings are in Appendices~\ref{app:detectors} and~\ref{app:controls}. Identical seeds do not guarantee bitwise agreement across environments.

Table~\ref{tab:compute} reports elapsed execution time from the run records. Times include data loading and evaluation inside each run and exclude scheduler waiting. CNN runs allocated at most 3.3\,GiB of GPU memory; official-test launchers requested 6--16 CPU cores and 80--256\,GB of host memory. The public CV2024 training and validation images occupy about 0.4\,GB.

Kvasir-Capsule runs timed only their training loops, 2.92 hours in total; the estimate in Table~\ref{tab:compute} adds evaluation and setup time, scaling each run's own training throughput by the evaluation-to-training cost ratio of the 30 official-test runs (median 0.89) and adding the primary runs' median per-run setup (9\,s), which gives 4.77 hours. The maximum ratio (1.21) and setup (30\,s) would give 5.57 hours and a project total of 119.6 hours. Their hardware was not recorded; they train DINOv2-L/14 with the batch sizes that ran on single 48\,GB GPUs in the local experiments. Image measurements are CPU-bound: timing the released implementation on 800 images and 200 pairs on one core of an AMD Ryzen~7 7800X3D and extrapolating gives about 1.2 core-hours for all 105,362 images and 53,464 pairs. Replay, prediction recomputation and participant re-scoring each finish within a minute on one CPU core. The full project used more compute than Table~\ref{tab:compute}: 60 additional CNN diagnostic fits (5.26 hours) and preliminary and failed runs that were not fully recorded.

\begin{table}[htbp]
\centering
\caption{Compute of the reported experiments. Minutes per run give the range of elapsed time for one run (the mean for the auxiliary controls); Hours give each family's total elapsed execution time. GPU counts are as described in the text. Auxiliary CV2024 controls comprise the threshold, label, source-omission and add-back runs and their Full, Removal and Source-only references. The Kvasir-Capsule total and the overall total are estimates (see text).}
\label{tab:compute}
\small
\begin{tabular}{@{}lrlrr@{}}
\toprule
Experiment family & Runs & GPU (runs) & Minutes per run & Hours \\
\midrule
Primary DINOv2 & 110 & A100 (110) & 10.7--32.2 & 22.23 \\
Local removal & 40 & A6000 (24), A100 (8), L40S (8) & 28.8--64.7 & 24.34 \\
Frozen-backbone CNN & 60 & A100 (60) & 3.0--8.6 & 5.32 \\
Official test & 30 & L40S (28), A100 (1), A6000 (1) & 7.8--20.9 & 6.06 \\
Auxiliary CV2024 controls & 154 & A100, A6000, L40S & 21.8 (mean) & 55.96 \\
DINOv2 feature extraction & 1 & RTX 4070 Ti SUPER (1) & 9.5 & 0.16 \\
Kvasir-Capsule & 28 & not recorded & --- & $\approx$4.8 \\
\midrule
Total & 423 & & & $\approx$118.8 \\
\bottomrule
\end{tabular}
\end{table}

\subsection{Numerical replay and implementation scope}
\label{app:numerical}
Install the repository's pinned CPU dependencies and run, from the repository root,
\begin{quote}\small
\verb|python replay.py --output replay.json|
\end{quote}
with a new output filename. The command re-aggregates the main training and cohort comparisons from recorded metrics, recomputes local endpoints and seed/prefix intervals from saved probabilities, and aggregates official-test confusion matrices and class AUCs. It also recounts the included relationship measurements and SSIM control summaries. Then
\begin{quote}\small
\verb|python recompute.py --output recomputed.json|
\end{quote}
re-scores all 450 primary and CNN seed/view records from the saved probabilities and checks them, with the contrasts in Table~\ref{tab:main_evidence}(b), against the recorded values; Tables~\ref{tab:training_pools}, \ref{tab:control_contrasts}, \ref{tab:training_views}, \ref{tab:class_recall} and~\ref{tab:cnn_final} follow from the same predictions, and Table~\ref{tab:source_metrics} can be computed from their source labels. \texttt{rescore.py} re-scores the 25 readable participant workbooks fetched from the upstream repository for Tables~\ref{tab:main_evidence}(d) and~\ref{tab:m7}--\ref{tab:rank_alignment}. \texttt{kvasir\_splits.py}, written for this release rather than taken from the original analysis, rechecks the official Kvasir-Capsule list properties of Appendix~\ref{app:auxiliary} from the upstream lists at the fixed revision. Official-test AUC replay averages stored class AUCs; it does not reconstruct their rankings from per-image scores. These commands verify saved outputs; they do not rerun training, inference or image measurement, and they do not regenerate the typeset tables, figures or paper.

\texttt{reuse.py} implements byte-confirmed SHA-256 groups, KVASIR prefix counts, independent pHash/dHash minima and $96\times96$ exact-pHash counterpart NCC, with optional selected-pair SSIM. The minimum-distance function does not select reference identities. \texttt{controls.py} constructs Full, Source-only, Removal, Random, Comp, Class-match-R and Ulcer replacement A, plus the three evaluation views. Both modules encode CV2024's four sources and ten classes; \texttt{controls.py} also requires the 66-row Ulcer sets of replacement A and a screen that flags every KVASIR training row. Another release requires replacing these constants. \texttt{analysis.py} provides pooled/equal-source scoring, paired-$t$ and seed-bootstrap contrasts, the local difference-in-differences and official-test aggregation. \texttt{example.py} uses synthetic inputs. The \texttt{training} directory contains the DINOv2 primary and local-removal recipes, the CNN runner and trainer, and the official-test producer, each with its recorded environment. The official-test producer is the archived file, the CNN trainer's executable code is unchanged, and the DINOv2 recipes matched their archived producers on reduced CPU inputs in trainable tensors, random-number state, losses and epoch metrics. Seeds and data order are fixed, but mixed-precision GPU training is not guaranteed to be bitwise reproducible, even on identical hardware; the DINOv2 Torch Hub revision and frozen-weight digest of the original runs were not recorded, and the official-test producer saved no checkpoints or per-image predictions.

The release omits local/distant sampling code, construction code for controls B--D, PDQ and feature/transform audits, reference-counterpart selection (Table~\ref{tab:ncc_bands}), the basename and joint-rule candidate searches (Tables~\ref{tab:internal_merged}--\ref{tab:internal_pixels}), matched-SSIM control construction, the $256\times256$ NCC implementation, video-prefix resampling (Table~\ref{tab:clusters}), the official-test byte scan, the external-release screens and code for the auxiliary CV2024 and Kvasir-Capsule training analyses. Retained outputs of the counterpart selection and candidate searches are included where available. The alternate-resolution result is included only as retained per-prefix counts. Kvasir-Capsule histories lack per-run hardware records and verified links to the code of each run.

New hash and pixel measurements require original images; hash-based candidate searches can use recorded measurements. Model training and inference additionally require the pretrained weights and a CUDA GPU; the official-test producer also reads the Kvasir-Capsule images and its \texttt{split\_1} list. Patient-level validation requires identifiers absent from the public metadata.

\subsection{Example: a training-pool contrast}
\label{app:record}
This example uses ReuScope's record structure for a training comparison and complements the relationship record in Section~\ref{sec:protocol}. Its ordered inputs and seed-level metric records are included in \texttt{data/results}.

\begin{description}
\item[Question and boundary.] How does Full-to-Removal training change public-validation BA?
\item[Inputs and metric.] Full/Removal contain 37,607/10,591 training rows. Their models share the 16,132-row validation list; BA averages recall over the same ten classes, and seeds 0--9 are paired.
\item[Estimate and uncertainty.] Mean BA difference: $\RemovalDelta$ (95\% paired-$t$: $\RemovalCI$). This pointwise interval reflects seed variation conditional on fixed lists and pipeline, without adjustment for analysis selection.
\item[Conclusion.] Removal has lower public-validation BA under this pipeline; the comparison does not isolate the contribution of leakage.
\end{description}

\subsection{Data, software and model terms}
The public implementation's \texttt{LICENSE} applies MIT to author-written code. Author-created measurements, predictions and derived results in the repository's \texttt{data} folder use CC BY 4.0, as stated in its license notice. Adapted CV2024 row metadata retain provider attribution and terms. Neither license grants rights to upstream images or weights. Organizer code is excluded because the examined repository lacks a repository-level LICENSE; scoring follows this paper's metric definitions, and participant workbooks are read from that repository rather than redistributed.

The examined CV2024 training/validation and test release records state CC BY 4.0~\citep{capsule_vision_trainval_figshare,capsule_vision_test_figshare}; the test record also specifies research-only use. The challenge paper states research-only use for the datasets collectively~\citep{capsule_vision_2024}. Among its component sources, the SEE-AI dataset states CC BY 4.0 and asks users to cite its article~\citep{yokote2024seeai}, KID provides its datasets for academic research upon registration~\citep{koulaouzidis2017kid}, and the AIIMS frames come from a private collection~\citep{goel2022dilated}; Kvasir-Capsule terms are given below. Kvasir-Capsule's article and repository state CC BY 4.0, while the repository also includes competition/commercial permission wording. The ISIC 2019 training release is provided under CC BY-NC 4.0, with source credit to BCN\_20000 (Department of Dermatology, Hospital Cl\'inic de Barcelona), HAM10000 (ViDIR Group, Department of Dermatology, Medical University of Vienna) and MSK (Anonymous). HyperKvasir is provided under CC BY 4.0; its repository also requires prior written permission for competition or commercial use.\footnote{HyperKvasir provider terms: \url{https://github.com/simula/hyper-kvasir}.} Users obtain the original images under the applicable provider terms.

NIH's dataset FAQ describes use as unrestricted and requests the original download link, NIH Clinical Center acknowledgment and the CVPR 2017 citation~\citep{wang2017chestxray}. We acknowledge the NIH Clinical Center as the provider of the ChestX-ray14 identifier lists.\footnote{\raggedright NIH FAQ, p.~1, Q04, accessed 27 October 2026: \url{https://nihcc.app.box.com/v/ChestXray-NIHCC/file/249502714403}.}

DINOv2 code and pretrained weights use Apache 2.0. PyTorch, TorchVision, NumPy, SciPy, pandas, scikit-learn and scikit-image use BSD-3-Clause, subject to file-specific notices; ImageHash uses BSD-2-Clause, and Pillow uses MIT-CMU. The \texttt{pdqhash} Python bindings use MIT; underlying PDQ uses BSD-3-Clause.\footnote{\raggedright DINOv2: \url{https://github.com/facebookresearch/dinov2}; PDQ bindings: \url{https://github.com/faustomorales/pdqhash-python}.}

TorchVision's ResNet-50 \texttt{IMAGENET1K\_V2} and ConvNeXt-Tiny \texttt{IMAGENET1K\_V1} checkpoints use ImageNet-1K pretraining~\citep{deng2009imagenet,russakovsky2015imagenet}. TorchVision notes that pretrained weights may carry training-data terms; ImageNet's database conditions specify noncommercial research and education.\footnote{\raggedright TorchVision pretrained-model notice: \url{https://docs.pytorch.org/vision/stable/models.html}; ImageNet database terms: \url{https://www.image-net.org/download.php}.} Model identifiers are specified in Appendix~\ref{app:controls}. Weights are obtained separately under provider terms, not redistributed or relicensed.

\paragraph{LLM assistance.}
LLMs assisted only with code and manuscript preparation; they are not components of ReuScope, the image models or the numerical and statistical calculations.

\end{document}